\documentclass[10pt,twocolumn,letterpaper]{article}
\usepackage[T1]{fontenc}
\usepackage[pagenumbers]{cvpr}

\DeclareRobustCommand{\methodname}{\mbox{%
  \textcolor[HTML]{C8503A}{D}\textcolor[HTML]{C8503B}{r}\textcolor[HTML]{C6513E}{e}%
  \textcolor[HTML]{C05448}{a}\textcolor[HTML]{B2595B}{m}\textcolor[HTML]{9B5E76}{c}%
  \textcolor[HTML]{7F6093}{a}\textcolor[HTML]{665FA9}{t}\textcolor[HTML]{565DB6}{c}%
  \textcolor[HTML]{4E5CBC}{h}\textcolor[HTML]{4B5BBE}{e}\textcolor[HTML]{4B5BBE}{r}}}
\newcommand{\method}{\methodname\xspace}
\usepackage[expansion=false]{microtype}
\usepackage{needspace}

\usepackage{fvextra}
\usepackage[skins]{tcolorbox}
\newenvironment{promptbox}[1]{%
  \VerbatimEnvironment
  \begin{tcolorbox}[enhanced,
    colback=black!4, colframe=black!35, boxrule=0.4pt, arc=1.5pt,
    left=3pt, right=3pt, top=1pt, bottom=1pt, before skip=6pt, after skip=6pt,
    title={#1}, fonttitle=\footnotesize\bfseries, coltitle=black, colbacktitle=black!12,
    toptitle=1pt, bottomtitle=1pt]%
  \begin{Verbatim}[fontsize=\scriptsize, breaklines=true, breaksymbolleft={}]}%
  {\end{Verbatim}\end{tcolorbox}}

\AtBeginDocument{%
  \expandafter\appto\csname normalsize \endcsname{%
    \setlength{\abovedisplayskip}{5pt plus 2pt minus 2pt}%
    \setlength{\belowdisplayskip}{5pt plus 2pt minus 2pt}}%
  \normalsize}

\usepackage{colortbl}  
\usepackage{multirow}  
\usepackage{fontawesome5}  
\definecolor{oursbg}{RGB}{230,239,250}  
\newcommand{\oursrow}{\rowcolor{oursbg}}  
\newcommand{\compacttable}{%
  \renewcommand{\arraystretch}{0.91}%
  \setlength{\aboverulesep}{0.35ex}%
  \setlength{\belowrulesep}{0.5ex}%
}

\newcommand{\fitwidth}[1]{\resizebox{\ifdim\width>\linewidth\linewidth\else\width\fi}{!}{#1}}
\newsavebox{\fillwidthbox}
\newlength{\mintabcolsep}
\newcommand{\fillwidth}[2]{%
  \begingroup
  \setlength{\tabcolsep}{0pt}%
  \sbox{\fillwidthbox}{#2}%
  \setlength{\tabcolsep}{\dimexpr(\linewidth-\wd\fillwidthbox)/(2*#1)\relax}%
  \ifdim\tabcolsep<\mintabcolsep
    \setlength{\tabcolsep}{\mintabcolsep}%
    \resizebox{\linewidth}{!}{#2}%
  \else
    \sbox{\fillwidthbox}{#2}%
    \setlength{\tabcolsep}{\dimexpr\tabcolsep+(\linewidth-\wd\fillwidthbox)/(2*#1)\relax}%
    #2%
  \fi
  \endgroup}
\definecolor{catprop}{HTML}{4C5BA8}       
\definecolor{catoffline}{HTML}{D9A13A}    
\definecolor{catonline}{HTML}{2A9D8F}     
\definecolor{basebg}{HTML}{EFEBE2}        
\definecolor{grouprule}{RGB}{185,192,200} 
\definecolor{trainedc}{RGB}{212,98,42}    
\definecolor{trainfreec}{RGB}{47,127,193} 
\DeclareRobustCommand{\catlegend}[1]{%
  \raisebox{0.15ex}{\resizebox{0.85ex}{0.85ex}{\textcolor{#1}{\faCircle}}}}
\newcommand{\tablemarker}[1]{\makebox[1em][c]{\smash{#1}}\hspace{0.4em}}
\newcommand{\catprop}{\tablemarker{\catlegend{catprop}}}
\newcommand{\catoffline}{\tablemarker{\catlegend{catoffline}}}
\newcommand{\catonline}{\tablemarker{\catlegend{catonline}}}
\newcommand{\catnone}{\tablemarker{}}
\newcommand{\baseicon}[1]{%
  \tablemarker{\raisebox{-0.2em}{\includegraphics[height=1em,width=1em,keepaspectratio]{figures/icons/#1.png}}}\ignorespaces}
\newcommand{\baserow}{\rowcolor{basebg}}
\newcommand{\tabgroupsep}{\arrayrulecolor{grouprule}\midrule\arrayrulecolor{black}}
\DeclareRobustCommand{\legendbox}[2]{%
  {\setlength{\fboxsep}{1.5pt}\raisebox{0pt}[0pt][0pt]{\colorbox{#1}{#2}}}}

\newcommand{\best}[1]{\textbf{#1}}
\newcommand{\second}[1]{\underline{#1}}

\definecolor{cvprblue}{rgb}{0.21,0.49,0.74}
\usepackage[breaklinks,colorlinks,allcolors=cvprblue]{hyperref}
\hypersetup{
  pdftitle={Dreamcatcher: Remembering What Surprises a World Model in Video Streams},
  pdfauthor={Jinxi He, Ce Zhang, Wenhe Liu, Jianshu Zhang, Katia P. Sycara, Yaqi Xie}
}

\title{\texorpdfstring{\methodname}{Dreamcatcher}: Remembering What Surprises a World Model in Video Streams}
\author{%
  \textbf{Jinxi He\textsuperscript{1} \quad Ce Zhang\textsuperscript{1} \quad
    Wenhe Liu\textsuperscript{2} \quad Jianshu Zhang\textsuperscript{3} \quad
    Katia P. Sycara\textsuperscript{1} \quad Yaqi Xie\textsuperscript{4}}\\[4pt]
  {\small \textsuperscript{1}Robotics Institute, Carnegie Mellon University \quad
          \textsuperscript{2}Language Technologies Institute, Carnegie Mellon University}\\
  {\small \textsuperscript{3}Northwestern University \quad
          \textsuperscript{4}University of Illinois at Urbana-Champaign}%
}

\begin{document}
\twocolumn[{%
  \maketitle
  \vspace{-12pt}
  {
  \centering
  \captionsetup{hypcap=false}
  \includegraphics[width=\textwidth]{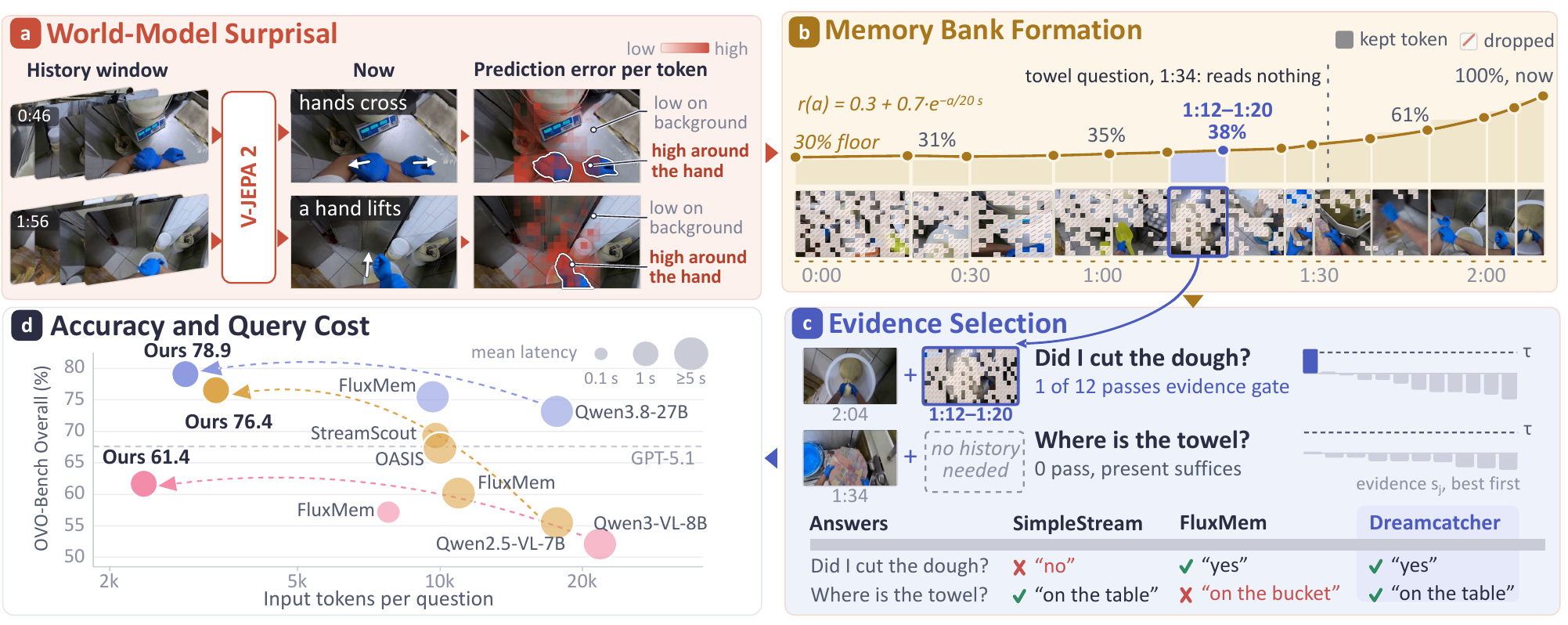}
  \vspace{-18pt}
  \captionof{figure}{\textbf{\method remembers what surprises a world model and reads history only when the current view falls short.}
  (a) A frozen V-JEPA~2 predicts incoming clip features from recent history.
  Surprisal maps highlight the crossing and lifting hands, with lower scores over the background.
  (b) The Online Budget Allocator segments the stream and stores each segment as native visual tokens of the frozen Video-LLM, favoring surprising tokens under a budget that decays with its age toward a 30\% floor.
  (c) The Evidence Gate admits one shortlisted segment for the dough question and none for the towel question.
  The table compares answers from SimpleStream, FluxMem, and \method, with check marks for correct ones.
  Unlike the other two methods, \method answers both correctly.
  (d) On OVO-Bench, \method achieves the highest overall accuracy among the compared training-free methods and reduces input tokens for answer generation by more than 80\% relative to its base model at 1\,fps.}
  \label{fig:teaser}
  \par
}

  \par\medskip
}]
\AtEndEnvironment{abstract}{%
  Our project page is at
  \href{https://jinxihe03.github.io/dreamcatcher/}{jinxihe03.github.io/dreamcatcher/}.%
}
\begin{abstract}
As live video streams grow longer, repeatedly encoding and processing their full histories becomes increasingly costly.
Answering questions as they arrive therefore calls for visual memory built in one pass before future questions are known.
Existing methods mainly keep tokens that differ from earlier ones or match surrogate queries fixed in advance.
Without modeling world dynamics, these criteria struggle to tell subtle surprising details from expected change.
We introduce \method, a training-free framework and, to our knowledge, the first to use predictive signals from an independently pretrained, frozen video world model for online memory formation in streaming video question answering.
Conditioned on recent history, the world model predicts incoming features using learned world priors.
Its normalized prediction errors yield temporal and spatial surprisal signals that guide online segmentation and the retention of a frozen Video-LLM's native visual tokens, balancing surprising details with broad visual coverage under budgets that decay with segment age.
At query time, an evidence gate admits a segment from a similarity-based shortlist only when the Video-LLM judges that it adds evidence for the question beyond the current view.
We evaluate \method on three streaming and three offline long-video benchmarks.
On OVO-Bench, \method achieves the highest overall accuracy among the compared training-free methods for each of four frozen Video-LLMs, reaching 78.9\% with Qwen3.8-27B.
With the three Qwen Video-LLMs, it also reduces average input tokens for answer generation by more than 80\% relative to their base models at 1\,fps.
\end{abstract}
\vspace{-10pt}

\section{Introduction}
\label{sec:intro}

Streaming video understanding requires a system to answer questions about current and past events using only the video it has seen so far~\cite{videollmonline,streamchat,streamingbench,ovobench}.
As new frames arrive, re-encoding this growing history for each question, or caching and attending to all of it, becomes increasingly expensive~\cite{flashvstream,timechatonline,rekv,infinipotv}.
This cost poses a substantial barrier to deploying always-on visual assistants in live streaming~\cite{livecc}, wearable devices~\cite{egolife}, and embodied robots~\cite{remembr}, where tasks such as episodic recall~\cite{ego4d,openeqa}, task guidance~\cite{holoassist}, and long-horizon navigation~\cite{smt} depend on events observed earlier.

\looseness=-1
While existing methods have primarily reduced this cost by discarding or merging visual tokens that resemble earlier ones~\cite{timechatonline,infinipotv,streamforest,fluxmem,vtc} or leave the Video-LLM's recurrent state unchanged~\cite{deltas}, how to selectively retain the evidence that later questions will depend on remains an open problem.
This choice must be made in a single pass, before any question is known.
With no question to guide them, these methods in effect predict that the stream will repeat itself and retain whatever departs from this prediction.
Other methods go a step further and predict with constant-velocity extrapolation or codec motion compensation~\cite{surge,codecguided}.
Still others anticipate questions with surrogate queries fixed in advance~\cite{savemem,streammem}.
However, these simple predictions and fixed queries capture little of world dynamics.
An action can change much of the frame yet be entirely expected, whereas a detail can change only slightly yet be surprising.
While a person chops vegetables, the knife and hands change many tokens in every frame, yet the motion repeats.
Such expected change can then fill the memory and crowd out the surprising details that later questions may ask about.

\looseness=-1
These limitations motivate our central question:
\textit{Can a predictor that has learned world dynamics tell surprising details from expected change?}
Recent video world models provide such a predictor~\cite{vjepa2}.
Conditioned on recent history, they predict the incoming content, and the prediction error measures how far the stream departs from what was expected.
Latent-space models such as V-JEPA~2 do so without generating frames.
They focus on predictable dynamics, such as the trajectory of a moving object, rather than unpredictable pixel-level detail, such as each leaf on a tree~\cite{vjepa,vjepa2,lecun2022path}.
Their prediction errors identify physically implausible events~\cite{garrido2025physics}, and related surprise signals have helped robot episodic memory~\cite{worthremembering} and continual spatial reasoning~\cite{cambrians}.
For a streaming memory, these errors could indicate, without any question, when and where the unexpected occurs, so that older content gives up its predictable tokens first.

However, how to exploit world-model surprisal to guide memory formation for a Video-LLM, and how to select the most helpful history for each question, remains a critical challenge.
Methods in other domains use similar surprisal scores to guide memory, but they either retrain the Video-LLM jointly with their own predictor~\cite{cambrians} or replace the world model's predictor with a statistical surrogate, leaving its predictions unused~\cite{worthremembering}.
To address this, inspired by how humans segment ongoing activity at moments of high prediction error~\cite{zacks2007} and let memories fade with time~\cite{ebbinghaus1913}, we introduce \method, a training-free framework built on a frozen, off-the-shelf V-JEPA~2.
\emph{World-Model Surprisal} normalizes the prediction errors over time and maps them onto the Video-LLM's visual tokens as block-level scores and spatial maps (\cref{fig:teaser}a).
The \emph{Online Budget Allocator} uses these scores to segment the stream and to keep the most surprising native visual tokens under budgets that decay with age, together with a share for broad coverage (\cref{fig:teaser}b).
Since similarity alone does not reveal which history helps, the \emph{Evidence Gate} asks the Video-LLM whether each shortlisted segment adds evidence beyond the current view and admits only those that do, possibly none (\cref{fig:teaser}c).

Our contributions can be summarized as follows:
\begin{itemize}[topsep=3pt,itemsep=3pt,partopsep=0pt]
  \item We introduce \emph{World-Model Surprisal}, which, to our knowledge, is the first to use predictive signals from an independently pretrained, frozen video world model to guide online memory formation for streaming video question answering.

  \item We develop \method, a training-free framework that integrates online memory formation with query-time evidence selection. Its \emph{Online Budget Allocator} maintains a memory bank of native visual tokens under budgets that decay with age, while its \emph{Evidence Gate} selects history judged to add evidence beyond the current view.

  \item Extensive experiments on three streaming and three offline long-video benchmarks demonstrate strong performance across multiple frozen Video-LLMs and substantial reductions in input tokens for answer generation.
\end{itemize}

\begin{figure*}[t]
  \centering
  \includegraphics[width=\textwidth]{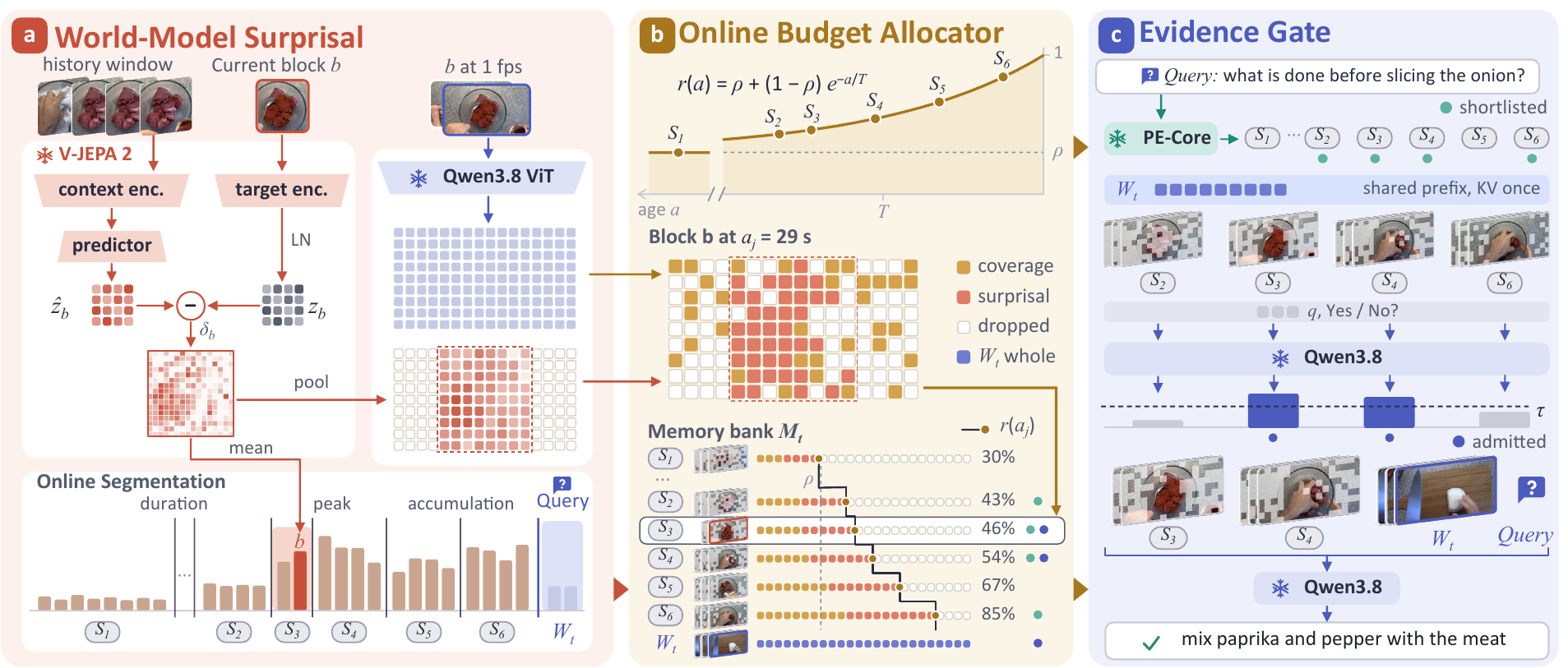}
  \vspace{-15pt}
  \caption{\textbf{Overview of \method.}
  (a) \emph{World-Model Surprisal} uses frozen V-JEPA~2 to predict block features from a sliding history window.
  Normalized prediction errors yield spatial scores pooled onto the reader's token grid and block-level scores for online segmentation.
  (b) The \emph{Online Budget Allocator} sets each segment's budget by a retention ratio $r(a)$ that decays with segment age toward the floor $\rho$. Ii combines coverage and surprisal to retain native visual tokens under budgets, while keeping the current anchor $W_t$ in full.
  (c) The \emph{Evidence Gate} shortlists segments by PE-Core~\cite{pecore} similarity to the question.
The reader scores each candidate for evidence beyond the current view in a separate pass, reusing the cached anchor $W_t$ as a shared prefix.
Admitted segments (here $S_3$ and $S_4$) meet threshold $\tau$ and join $W_t$ for answering.
Snowflakes denote frozen models.
  Snowflakes denote frozen models.}
  \label{fig:method}
  \vspace{-8pt}
\end{figure*}

\section{Related Work}
\label{sec:related}

\smallskip
\noindent\textbf{Streaming video understanding.}
Streaming video understanding requires answering questions over continuous video streams, spanning real-time perception, backward tracing, and proactive response~\cite{ovobench,streamingbench,streamchat,ovbench}.
Some approaches train models specifically for streaming~\cite{videollmonline,dispider,streamready}, while others equip frozen Video-LLMs with external memory, including full or compressed key--value caches~\cite{rekv,infinipotv,streammem,hermes}, visual features and textual context~\cite{streamchat}, or frames retrieved for on-demand processing~\cite{streamscout,oasis}.
Strong recent-window baselines~\cite{simplestreaming} highlight the need to determine when historical evidence helps.
Existing methods make this decision for the question as a whole, using question cues~\cite{shallowstream}, answer uncertainty~\cite{weavetime}, query--recent-view affinity~\cite{savemem}, or model-initiated requests for more evidence~\cite{streamscout,oasis,videoagent,videoseek}.
Retrieval-augmented approaches verify individual retrieved items, but against the question alone~\cite{sevila,vgent}.
\method instead gives a frozen Video-LLM a training-free memory of its own visual tokens, retained online under world-model surprisal, and reads it through an Evidence Gate that judges each shortlisted segment by the evidence it adds beyond the current view, so a question may read several segments or none. \looseness=-1

\smallskip
\noindent\textbf{Visual token selection and retention.}
Offline Video-LLMs prune or merge visual tokens, often guided by encoder, reader, or query attention~\cite{longvu,dycoke}.
Online memory formation requires selecting information before future questions are known.
Existing methods merge redundant tokens~\cite{moviechat,flashvstream}, prune by temporal similarity or frame residuals~\cite{timechatonline,stc,vtc}, or defer selection until query time~\cite{querystream}.
Others organize memory around entity or feature changes~\cite{memo,selectstream,curvestream}.
For long-term retention, methods use fixed memory tiers, favor recent content, or model temporal forgetting~\cite{fluxmem,streamforest,streambridge,ovbench,freshmem,hermes,streamchat}, sometimes reserving tokens for uniform coverage~\cite{tokenpruning,savemem}.
Most of these criteria measure how much the stream changes, not whether the change was expected.
The \emph{Online Budget Allocator} in \method instead ranks native visual tokens by world-model surprisal, keeps a coverage share for predictable content, and lets budgets decay with segment age, following the forgetting curve of human memory~\cite{ebbinghaus1913}.

\smallskip
\looseness=-1
\noindent\textbf{Predictive signals for video memory.}
Video compression and offline long-video understanding methods estimate surprise through codec signals~\cite{codecguided,codecsight}, predictive coding~\cite{adacodec}, temporal extrapolation~\cite{swiftsampling,surge}, or changes in a Video-LLM's beliefs~\cite{spike}.
Related signals guide event segmentation in language-model memory~\cite{emllm} and episodic retention in robotics.
Worth Remembering estimates surprise with a sliding-window Gaussian over V-JEPA~2 embeddings~\cite{worthremembering}, while Cambrian-S jointly trains a predictor with the reader for streaming memory~\cite{cambrians}.
The prediction errors of V-JEPA have also been used to measure surprise in tests of intuitive physics~\cite{garrido2025physics}.
Unlike hand-crafted or jointly trained predictors, \emph{World-Model Surprisal} takes spatially resolved errors directly from an independently pretrained, frozen V-JEPA~2 predictor and, without seeing any question, uses them to decide which native visual tokens a frozen Video-LLM retains.

\section{Method}
\label{sec:method}

\method forms visual memory as a stream arrives and selects evidence from that memory when a question is asked (\cref{fig:method}).
\emph{World-Model Surprisal} scores incoming content, the \emph{Online Budget Allocator} turns these scores into segments of retained native visual tokens, and the \emph{Evidence Gate} selects history at query time.
The world model, retrieval encoder, and Video-LLM remain frozen throughout.
Appendix~\ref{sec:supp_impl} details each component.

\subsection{Preliminaries}
\label{sec:setup}

\smallskip
\noindent\textbf{Streaming video question answering.}
A video stream delivers a potentially unbounded sequence of frames $\{f_1,f_2,\dots\}$ in temporal order.
A question $q$ issued at time $t$ must be answered using only the observed prefix $V_{\le t}$.
As the prefix grows, the system maintains a compact causal memory $\mathcal{M}_t$ through one-pass processing.
Memory updates cannot depend on future questions, and discarded historical frames cannot be revisited.

\smallskip
\looseness=-1
\noindent\textbf{Visual memory.}
The reader is a frozen Video-LLM whose vision encoder maps each short temporal \emph{block} $b$ to native visual tokens $X_b=\{x_{b,i}\}_{i=1}^{N_b}$.
Consecutive blocks form a \emph{segment} $S_j$; its selected token subset at time $t$ is denoted by $\widetilde{X}_j(t)$.
Blocks are the fixed units of visual encoding, whereas segments adapt to the stream and serve as the units of retention and retrieval.
The \emph{visual anchor} $W_t$ comprises the latest portion of the memory bank, whose visual tokens are retained in full.
Only segments that end before $W_t$ begins can be retrieved.

\subsection{World-Model Surprisal}
\label{sec:surprisal}
\label{sec:write}

A pretrained world model predicts incoming content from the observed history, and we call the normalized error of this prediction in the model's feature space \emph{surprisal}.
It requires no query and supplies a temporal score for segmentation and a spatial map for token selection.

\smallskip
\noindent\textbf{Prediction-based scoring.}
We use frozen V-JEPA~2, which consists of a context encoder, a predictor, and a target encoder~\cite{vjepa2}.
For each incoming block $b$, the context encoder processes a fixed-length sliding window of preceding observations, the predictor estimates the target-encoder features of $b$, and the target encoder supplies the observed features.
For target spatiotemporal patch $p$, the residual between the predictor output and the layer-normalized target is $\delta_{b,p}=\hat{z}_{b,p}-\operatorname{LN}(z_{b,p})\in\mathbb{R}^{D}$.
To make channels with different error scales comparable and to track changes in the stream's error scale, we define
\begin{equation}
  u_{b,p} = \frac{1}{D} \sum_{d=1}^{D} \min\!\left( \frac{|\delta_{b,p,d}|}{\max(\sigma_{b,d}, \sigma_{\min})},\, \kappa \right),
  \label{eq:residual}
\end{equation}
where $\sigma_{b,d}$ is a robust scale estimated from earlier blocks' channel errors, $\sigma_{\min}$ prevents unstable normalization, and $\kappa$ limits the influence of extreme errors.

\smallskip
\noindent\textbf{Temporal and spatial scores.}
Averaging over all target patches gives the block score $\bar{u}_b$, which indicates when prediction difficulty rises but not which details within a block merit storage.
For token selection, we average patch scores across temporal positions and pool the resulting map onto the reader's token grid by area overlap, obtaining $u_{b,i}$ (\cref{fig:method}a).
Only the scores are transferred, and the retained representations remain the reader's own visual tokens.

\begin{table*}[t]
  \centering
  \caption{\textbf{Performance comparison on OVO-Bench.}
    Colored dots mark \catlegend{catprop}\,proprietary, \catlegend{catoffline}\,open-source offline, and \catlegend{catonline}\,open-source streaming models.
    Each base-model block lists the \legendbox{basebg}{base model}, training-free methods, and \legendbox{oursbg}{\method}.
    Within each block, the best score in each column is in \textbf{bold} and the second best is \underline{underlined}.
    Task definitions appear in Appendix~\ref{sec:supp_impl}.}
  \label{tab:ovo_comparison}
  \vspace{-8pt}
  \footnotesize
  \compacttable
  \fillwidth{12}{%
  \begin{tabular}{l@{\hspace{4pt}}l@{\hspace{4pt}} ccccccc cccc c}
    \toprule
    \multirow{2}{*}{\textbf{Model}} & \multirow{2}{*}{\textbf{Input}} & \multicolumn{7}{c}{\textbf{Real-Time Visual Perception}} & \multicolumn{4}{c}{\textbf{Backward Tracing}} & \multirow{2}{*}{\textbf{Overall}} \\
    \cmidrule(lr){3-9} \cmidrule(lr){10-13}
    & & OCR & ACR & ATR & STU & FPD & OJR & Avg & EPM & ASI & HLD & Avg & \\
    \midrule

    \catprop GPT-4o & 64 frames & 69.8 & 64.2 & 71.6 & 51.1 & 70.3 & 59.8 & 64.5 & 57.9 & 75.7 & 48.7 & 60.8 & 62.6 \\
    \catprop GPT-5-mini & 64 frames & 81.2 & 67.9 & 81.9 & 66.3 & 73.3 & 66.3 & 72.8 & 66.7 & 75.7 & 16.1 & 52.8 & 62.8 \\
    \catprop GPT-5.1 & 64 frames & 83.2 & 78.9 & 77.6 & 67.4 & 79.2 & 76.6 & 77.2 & 62.0 & 73.7 & 38.7 & 58.1 & 67.6 \\
    \catoffline Qwen2-VL-72B & 64 frames & 65.8 & 60.6 & 69.8 & 51.7 & 69.3 & 54.4 & 61.9 & 52.5 & 60.8 & 57.5 & 57.0 & 59.4 \\
    \catoffline LLaVA-Video-7B & 64 frames & 69.1 & 58.7 & 68.8 & 49.4 & 74.3 & 59.8 & 63.4 & 56.2 & 57.4 & 7.5 & 40.4 & 51.9 \\
    \catoffline InternVL2-8B & 64 frames & 67.1 & 60.6 & 63.8 & 46.1 & 68.3 & 56.5 & 60.4 & 48.2 & 57.4 & 24.7 & 43.4 & 51.9 \\
    \catonline VideoLLM-online-8B~\cite{videollmonline} & 2\,fps & 8.1 & 23.9 & 12.1 & 14.0 & 45.5 & 21.2 & 20.8 & 22.2 & 18.8 & 12.2 & 17.7 & 19.3 \\
    \catonline Flash-VStream-7B~\cite{flashvstream} & 1\,fps & 24.2 & 29.4 & 28.5 & 33.7 & 25.7 & 28.8 & 28.4 & 39.1 & 37.2 & 5.9 & 27.4 & 27.9 \\
    \catonline Dispider-7B~\cite{dispider} & 1\,fps & 57.7 & 49.5 & 62.1 & 44.9 & 61.4 & 51.6 & 54.5 & 48.5 & 55.4 & 4.3 & 36.1 & 45.3 \\
    \catonline StreamForest-7B~\cite{streamforest} & 1\,fps & 68.5 & 53.2 & 71.6 & 47.8 & 65.4 & 60.9 & 61.2 & 58.9 & 64.9 & 32.3 & 52.0 & 56.6 \\
    \catonline TimeChat-Online-7B~\cite{timechatonline} & 1\,fps & 74.5 & 48.6 & 68.1 & 48.3 & 69.3 & 59.8 & 61.4 & 56.9 & 64.9 & 11.8 & 44.5 & 53.0 \\

    \tabgroupsep
    \baserow \baseicon{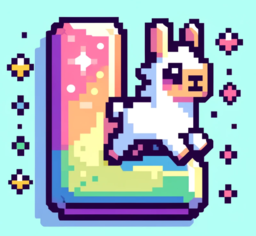} LLaVA-OneVision-7B & 64 frames & 66.4 & 57.8 & 73.3 & 53.4 & 71.3 & 62.0 & 64.0 & 54.2 & 55.4 & 21.5 & 43.7 & 53.9 \\
    \catnone HERMES~\cite{hermes} & 0.5\,fps & 72.5 & 62.4 & \second{74.1} & 50.6 & 73.3 & 65.2 & 66.3 & \best{60.6} & \best{61.5} & 28.5 & \second{50.2} & 58.3 \\
    \catnone SimpleStream~\cite{simplestreaming} & 4 frames & 82.7 & \second{65.1} & 72.2 & \second{55.4} & 71.2 & 71.0 & \second{69.6} & 50.8 & 52.1 & \second{41.7} & 48.2 & \second{58.9} \\
    \catnone ReKV~\cite{rekv} & 0.5\,fps & 73.8 & 56.0 & \second{74.1} & 51.7 & 70.3 & 60.3 & 64.4 & 54.2 & 57.4 & 28.5 & 46.7 & 55.5 \\
    \catnone STC~\cite{stc} & 0.5\,fps & 64.4 & 59.6 & 68.1 & 48.9 & 65.4 & 56.5 & 60.5 & 51.2 & 52.0 & 33.3 & 45.5 & 53.0 \\
    \catnone FluxMem~\cite{fluxmem} & 1\,fps & \second{83.2} & 62.4 & 73.3 & 51.1 & \second{74.3} & \second{71.7} & 69.3 & \second{56.6} & \second{58.1} & 22.0 & 45.6 & 57.5 \\
    \oursrow \catnone \textbf{\method{} (Ours)} & 1\,fps & \best{85.9} & \best{68.8} & \best{75.0} & \best{56.7} & \best{75.2} & \best{76.6} & \best{73.1} & 55.2 & 54.1 & \best{44.1} & \best{51.1} & \best{62.1} \\

    \tabgroupsep
    \baserow \baseicon{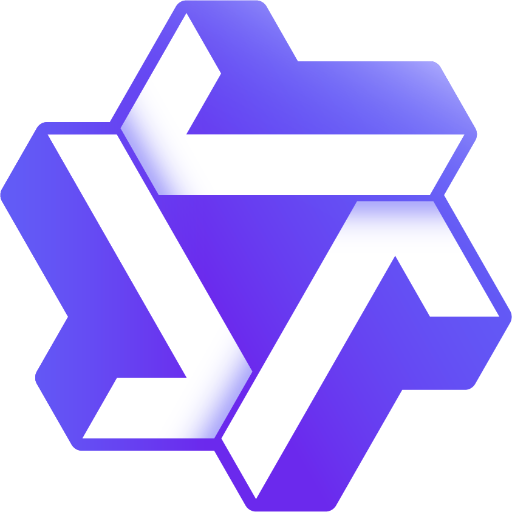} Qwen2.5-VL-7B & 1\,fps & 72.5 & 51.4 & 65.5 & 49.4 & 68.3 & 56.5 & 60.6 & \second{51.5} & 61.5 & 17.7 & 43.6 & 52.1 \\
    \catnone SimpleStream~\cite{simplestreaming}  & 4 frames & \second{85.2} & \best{71.6} & \second{72.4} & \second{57.9} & 69.3 & \second{72.3} & \second{71.4} & 46.1 & 56.8 & \best{38.7} & \best{47.2} & \second{59.3} \\
    \catnone QueryStream~\cite{querystream} & 1\,fps & 75.2 & 49.5 & 69.8 & 50.0 & \second{71.3} & 62.5 & 63.1 & \best{56.9} & \best{65.5} & 12.4 & \second{44.9} & 54.0 \\
    \catnone FluxMem~\cite{fluxmem} & 1\,fps & 81.2 & \second{59.6} & 70.7 & 53.4 & \best{75.3} & 63.0 & 67.2 & 48.5 & \second{64.2} & 29.0 & \best{47.2} & 57.2 \\
    \oursrow \catnone \textbf{\method{} (Ours)} & 1\,fps & \best{91.9} & \best{71.6} & \best{81.0} & \best{62.9} & \second{71.3} & \best{75.5} & \best{75.7} & 49.5 & 55.4 & \second{36.6} & \best{47.2} & \best{61.4} \\

    \tabgroupsep
    \baserow \baseicon{qwen} Qwen3-VL-8B & 1\,fps & 78.5 & 63.3 & 73.3 & 57.3 & 69.3 & 59.8 & 66.9 & 55.9 & 66.2 & 10.2 & 44.1 & 55.5 \\
    \catnone StreamScout~\cite{streamscout} & 4/12/24 frames & 91.3 & 82.6 & \best{82.8} & 64.0 & 75.3 & 75.0 & 78.5 & 55.2 & 61.5 & \second{63.4} & \second{60.1} & 69.3 \\
    \catnone HERMES~\cite{hermes} & 2\,fps & 87.9 & 74.3 & 79.3 & 56.2 & 70.3 & 71.7 & 73.3 & 55.6 & 67.6 & 24.7 & 49.3 & 61.3 \\
    \catnone SimpleStream~\cite{simplestreaming} & 4 frames & \best{94.0} & \best{85.3} & \best{82.8} & 65.7 & \best{77.2} & \best{83.2} & \best{81.4} & 51.9 & 58.1 & 52.1 & 54.0 & 67.7 \\
    \catnone OASIS~\cite{oasis} & 1--2\,fps & 92.0 & 80.7 & \second{81.0} & 67.4 & 67.3 & 79.9 & 78.1 & \best{62.0} & 60.1 & 47.3 & 56.5 & 67.3 \\
    \catnone ShallowStream~\cite{shallowstream} & 1\,fps & 92.0 & \second{83.5} & \best{82.8} & \second{71.9} & 75.3 & 79.9 & 80.9 & 52.5 & \best{72.3} & 49.5 & 58.1 & \second{69.5} \\
    \catnone FluxMem~\cite{fluxmem} & 1\,fps & 85.9 & 71.6 & 73.3 & 57.9 & 71.3 & 66.8 & 71.1 & 58.9 & 66.2 & 22.6 & 49.2 & 60.2 \\
    \oursrow \catnone \textbf{\method{} (Ours)} & 1\,fps & \second{93.3} & \second{83.5} & 78.4 & \best{72.5} & \second{76.2} & \second{82.1} & \second{81.0} & \second{61.3} & \second{71.6} & \best{82.8} & \best{71.9} & \best{76.4} \\

    \tabgroupsep
    \baserow \baseicon{qwen} Qwen3.8-27B~\cite{qwen38} & 1\,fps & 79.9 & 60.6 & \best{81.9} & 58.4 & 76.2 & 66.3 & 70.5 & \second{65.3} & \best{68.2} & 94.1 & \second{75.9} & 73.2 \\
    \catnone SimpleStream~\cite{simplestreaming}  & 4 frames & \second{94.6} & \best{78.9} & \second{80.2} & \second{68.5} & \second{79.2} & \second{82.1} & \best{80.6} & 65.0 & 53.4 & \best{96.8} & 71.7 & \second{76.1} \\
    \catnone FluxMem~\cite{fluxmem} & 1\,fps & 90.6 & \best{78.9} & \second{80.2} & \best{70.8} & 77.2 & \best{82.9} & \second{80.1} & 64.7 & \best{68.2} & 79.8 & 70.9 & 75.5 \\
    \oursrow \catnone \textbf{\method{} (Ours)} & 1\,fps & \best{95.3} & \second{75.2} & \second{80.2} & 67.4 & \best{80.2} & 80.4 & 79.8 & \best{72.1} & \second{66.2} & \second{95.7} & \best{78.0} & \best{78.9} \\
    \bottomrule
  \end{tabular}}
  \vspace{-10pt}
\end{table*}

\subsection{Online Budget Allocator}
\label{sec:allocator}

The allocator groups incoming blocks into segments, assigns budgets that decay with segment age, and selects native visual tokens within each budget.

\smallskip
\noindent\textbf{Online segmentation.}
Block-level surprisal scores organize the stream into segments (\cref{fig:method}a).
A local peak marks a sharp departure from recent predictions, accumulation captures sustained prediction difficulty, and a duration limit bounds segments when surprisal stays low or is unavailable.
With $S$ the open segment and $\zeta_b$ the robust $z$-score of $\bar{u}_b$ relative to preceding blocks, the three criteria are $\bar{u}_{b-1}<\bar{u}_b\ge\bar{u}_{b+1}$ with $\zeta_b\ge\zeta_0$ (peak), $\sum_{b'\in S}\bar{u}_{b'}\ge U_{\max}$ (accumulation), and $\operatorname{dur}(S)\ge L_{\max}$ (duration).
A peak places a boundary at the start of its block, the other two criteria close the open segment, and every boundary leaves a segment of at least $L_{\min}$.
A question at time $t$ also places a boundary at $\operatorname{start}(W_t)$, exempt from this minimum and independent of what the question asks.

\smallskip
\noindent\textbf{Age-dependent budgets.}
Recent observations retain more detail, while older segments approach a minimum retention ratio.
For a segment ending at $e_j$, its age is $a_j=t-e_j$ and its retention ratio is
\begin{equation}
  r(a) = \rho + (1 - \rho)\, e^{-a/T},
  \label{eq:retention}
\end{equation}
where $\rho$ sets the minimum retention ratio and $T$ the decay timescale.
With $N_j=\sum_{b\in S_j}N_b$, the token budget is $B_j(t)=\operatorname{round}(r(a_j)N_j)$.
The nonzero floor keeps older observations accessible after their initial allocations decay.
Each block is encoded once, its coverage order and surprisal ranking are reused as its budget shrinks, and a token is deleted once no later budget can select it.

\smallskip
\noindent\textbf{Coverage and surprisal shares.}
Each budget is split into a coverage share and a surprisal share (\cref{fig:method}b).
The \emph{coverage share}, a fraction $\alpha$ of the budget, assigns each block $\operatorname{round}(\alpha r(a_j)N_b)$ tokens, chosen by farthest-point sampling on cosine similarity over the full token grid.
Every dropped token thus stays similar to a retained one, which protects predictable details that an unforeseen question may need.
The \emph{surprisal share} fills the rest of the budget with the unselected scored tokens of highest $u_{b,i}$ across the segment, so poorly predicted blocks receive more tokens.

\subsection{Evidence Gate}
\label{sec:gate}
\label{sec:read}

\looseness=-1
The usefulness of history depends on both the question and the evidence already present in $W_t$.
A semantically matching segment can repeat visible content or show a related event without supplying the needed evidence.
The gate therefore shortlists candidates by similarity and then asks the reader whether each adds evidence given the current anchor (\cref{fig:method}c).

\smallskip
\noindent\textbf{Candidate shortlisting.}
Each segment stores a retrieval key $\mathbf{v}_j$, the normalized mean of its sampled-frame embeddings from a frozen image--text encoder~\cite{pecore}.
At query time, each eligible segment $S_j$ receives the score $c_j = \hat{m}_j + \beta \exp(-g_j / T_{\mathrm{r}})$, where $\hat{m}_j$ is the standardized similarity of $\mathbf{v}_j$ to the question and, with a lower weight, its options, $g_j$ is the gap from the segment's end to the start of $W_t$, and $\beta$ and $T_{\mathrm{r}}$ set the recency preference.
The shortlist $\mathcal{C}_t$ holds the $k$ highest-scoring segments.

\smallskip
\noindent\textbf{Evidence scoring and admission.}
For each $j\in\mathcal{C}_t$, the reader receives $W_t$, $\widetilde{X}_j(t)$, and $q$ with any answer options, and judges whether the candidate adds evidence beyond the current view, including evidence for part of the question.
The score is the Yes--No logit difference at the first answer position, $s_j = \ell_{\text{Yes}}(q,W_t,\widetilde{X}_j(t)) - \ell_{\text{No}}(q,W_t,\widetilde{X}_j(t))$.
Candidates are scored independently against the same cached anchor, prefilled once; each pass processes only the candidate and question and generates no tokens.
The gate admits $\mathcal{A}_t = \operatorname{Top}_{M}(\{j\in\mathcal{C}_t:s_j\ge\tau\};\,s)$, selecting at most $M$ candidates with scores at least $\tau$ in descending score order.
Admitted segments precede $W_t$ in temporal order; if none is admitted, the reader answers from $W_t$ alone.

\begin{table*}[t]
  \centering
  \caption{\textbf{Performance comparison on StreamingBench and StreamBench.}
    Colored dots mark \catlegend{catprop}\,proprietary, \catlegend{catoffline}\,open-source offline, and \catlegend{catonline}\,open-source streaming models.
    Each base-model block lists the \legendbox{basebg}{base model}, training-free methods, and \legendbox{oursbg}{\method}.
    Avg weights task scores by their question counts, and a dash marks an unreported score.
    Within each base-model block, the best score in each column is in \textbf{bold} and the second best is \underline{underlined}.
    Task definitions appear in Appendix~\ref{sec:supp_impl}.}
  \label{tab:comparison_stream}
  \vspace{-10pt}
  \footnotesize
  \compacttable
  \fillwidth{14}{%
  \begin{tabular}{ll ccccc ccccccc}
    \toprule
    \multirow{2}{*}{\textbf{Model}} & \multirow{2}{*}{\textbf{Input}} & \multicolumn{5}{c}{\textbf{StreamingBench}~\cite{streamingbench}} & \multicolumn{7}{c}{\textbf{StreamBench}~\cite{streamchat}} \\
    \cmidrule(lr){3-7} \cmidrule(lr){8-14}
    & & RTVU & ACU & MCU & SQA & Avg & OS & LM & SM & CI & KG & SF & Avg \\
    \midrule
    \catprop GPT-4o & 64 frames & 73.3 & 41.2 & 38.4 & 32.8 & 65.0 & 60.5 & 61.2 & 64.4 & 72.3 & 93.9 & 74.7 & 71.0 \\
    \catprop GPT-5-mini & 64 frames & 82.6 & 46.0 & 52.8 & 54.4 & 75.3 & 68.6 & 75.6 & 73.2 & 79.8 & 64.4 & 79.8 & 73.6 \\
    \catprop GPT-5.1 & 64 frames & 83.5 & 49.6 & 51.2 & 52.4 & 76.0 & 67.2 & 67.0 & 72.0 & 79.5 & 66.8 & 82.1 & 72.4 \\
    \catoffline LongVA & 8 frames & -- & -- & -- & -- & -- & 41.1 & 47.4 & 57.6 & 59.8 & 80.7 & 66.1 & 58.7 \\
    \catoffline InternVL2-8B & 16 frames & 63.7 & 32.0 & 31.2 & 32.3 & 56.3 & 38.5 & 46.6 & 50.9 & 67.6 & 81.0 & 62.2 & 57.6 \\
    \catoffline Qwen2-VL-7B & 0.2--1\,fps & 69.0 & 31.2 & 26.0 & 39.6 & 60.5 & -- & -- & -- & -- & -- & -- & -- \\
    \catoffline Video-LLaMA2-7B & 32 frames & 49.5 & 24.8 & 26.8 & 18.7 & 43.5 & -- & -- & -- & -- & -- & -- & -- \\
    \catonline MovieChat~\cite{moviechat} & 32 frames & -- & -- & -- & -- & -- & 18.6 & 20.4 & 26.5 & 42.3 & 67.2 & 35.8 & 34.9 \\
    \catonline VideoLLM-online-8B~\cite{videollmonline} & 5\,fps & -- & -- & -- & -- & -- & 41.4 & 48.8 & 52.9 & 62.7 & 69.2 & 64.1 & 56.4 \\
    \catonline Flash-VStream-7B~\cite{flashvstream} & 1\,fps & -- & -- & -- & -- & -- & 37.1 & 44.5 & 48.6 & 58.1 & 66.4 & 59.2 & 52.2 \\
    \catonline StreamChat~\cite{streamchat} & 15\,fps & 60.1 & 28.8 & 22.4 & 32.0 & 52.6 & 40.7 & 43.6 & 47.1 & 63.0 & 89.1 & 73.8 & 59.2 \\

    \tabgroupsep
    \baserow \baseicon{qwen} Qwen2.5-VL-7B & 1\,fps & 74.0 & 28.0 & 26.0 & 40.8 & 64.2 & 34.6 & \best{47.9} & 44.8 & 60.7 & 63.1 & 65.1 & 52.7 \\
    \catnone StreamScout~\cite{streamscout} & 4/12/24 frames & \second{77.7} & \best{58.0} & 39.2 & \second{53.2} & \second{71.3} & 34.8 & 32.0 & 43.5 & 41.4 & \best{83.6} & 63.8 & 49.7 \\
    \catnone OASIS~\cite{oasis} & 48/80 frames & 70.6 & 38.8 & 36.0 & 46.0 & 63.6 & 35.9 & 39.4 & 44.1 & 54.5 & 76.9 & 64.9 & 52.4 \\
    \catnone SimpleStream~\cite{simplestreaming} & 4 frames & 76.5 & 56.4 & \second{48.8} & 49.2 & 70.7 & 26.1 & 27.4 & 42.2 & 49.2 & 79.2 & 67.2 & 48.4 \\
    \catnone FluxMem~\cite{fluxmem} & 1\,fps & 76.3 & 32.4 & 30.8 & \best{53.6} & 67.6 & \best{38.8} & \best{47.9} & \second{46.9} & \best{66.3} & 69.5 & \second{72.8} & \second{56.8} \\
    \oursrow \catnone \textbf{\method{} (Ours)} & 1\,fps & \best{80.8} & \second{57.2} & \best{56.0} & 52.4 & \best{74.9} & \second{38.7} & \second{44.5} & \best{49.2} & \second{65.2} & \second{80.1} & \best{74.8} & \best{58.5} \\

    \tabgroupsep
    \baserow \baseicon{qwen} Qwen3-VL-8B~\cite{qwen3vl} & 1\,fps & 76.7 & 37.6 & 34.8 & 50.0 & 68.4 & 38.8 & 50.2 & 53.4 & 74.7 & 64.4 & \second{71.5} & 58.7 \\
    \catnone StreamScout~\cite{streamscout} & 4/12/24 frames & \second{80.3} & 48.0 & 42.0 & 42.0 & 71.9 & \best{52.5} & 51.8 & \best{61.4} & 71.4 & 84.6 & 70.8 & \second{65.3} \\
    \catnone SimpleStream~\cite{simplestreaming} & 4 frames & 77.9 & \second{58.8} & \second{56.8} & \second{54.4} & \second{73.0} & 25.8 & 24.1 & 44.2 & 63.3 & 61.1 & 60.9 & 46.5 \\
    \catnone OASIS~\cite{oasis} & 1--2\,fps & 78.2 & 42.7 & 49.6 & 48.4 & 71.0 & 47.0 & \second{53.6} & 54.3 & 71.7 & \second{86.1} & 67.6 & 63.1 \\
    \catnone FluxMem~\cite{fluxmem} & 1\,fps & 79.8 & 45.2 & 36.8 & 52.8 & 71.7 & 45.1 & 52.4 & 57.6 & \second{75.4} & 71.8 & 70.1 & 62.0 \\
    \oursrow \catnone \textbf{\method{} (Ours)} & 1\,fps & \best{84.6} & \best{60.0} & \best{60.0} & \best{60.8} & \best{78.9} & \second{47.5} & \best{55.8} & \second{58.7} & \best{78.1} & \best{87.6} & \best{75.5} & \best{67.0} \\

    \tabgroupsep
    \baserow \baseicon{qwen} Qwen3.8-27B~\cite{qwen38} & 1\,fps & \second{80.1} & 44.8 & 34.0 & 43.6 & 71.0 & 44.8 & \second{58.1} & 57.2 & 76.1 & 75.8 & 73.5 & 64.1 \\
    \catnone SimpleStream~\cite{simplestreaming} & 4 frames & 78.1 & \second{62.8} & \second{54.0} & 49.2 & 72.8 & 36.1 & 31.0 & 46.9 & 67.7 & 78.9 & 69.9 & 54.8 \\
    \catnone FluxMem~\cite{fluxmem} & 1\,fps & 79.2 & 55.2 & 47.2 & \second{54.8} & \second{73.0} & \second{53.8} & 54.8 & \second{64.8} & \second{77.7} & \best{82.5} & \second{74.9} & \second{68.0} \\
    \oursrow \catnone \textbf{\method{} (Ours)} & 1\,fps & \best{85.1} & \best{64.0} & \best{59.2} & \best{61.6} & \best{79.7} & \best{66.6} & \best{70.3} & \best{69.6} & \best{85.2} & \second{81.2} & \best{78.8} & \best{75.1} \\
    \bottomrule
  \end{tabular}}
  \vspace{-8pt}
\end{table*}


\section{Experiments}
\label{sec:exp}

\looseness=-1
We evaluate whether world-model-informed memory preserves the evidence needed by questions that are unknown during stream processing, using three streaming and three offline long-video benchmarks and four frozen Video-LLMs.
We then analyze efficiency, show qualitative examples, and ablate the selection signal, the retention schedule, and the Evidence Gate.

\subsection{Experimental setup}
\label{sec:exp_setup}

\smallskip
\noindent\textbf{Benchmarks and metrics.}
On OVO-Bench~\cite{ovobench}, we evaluate 1,468 questions from six real-time perception and three backward-tracing tasks and report the mean of the two group averages as Overall.
StreamingBench~\cite{streamingbench} and the open-ended StreamBench~\cite{streamchat} questions are answered at their annotated times from the observed prefix, and Video-MME~\cite{videomme}, MLVU~\cite{mlvu}, and LongVideoBench~\cite{longvideobench} are replayed as streams and answered at the end.

\smallskip
\noindent\textbf{Readers and baselines.}
We use LLaVA-OneVision-7B, Qwen2.5-VL-7B, Qwen3-VL-8B~\cite{qwen3vl}, and Qwen3.8-27B~\cite{qwen38} on OVO-Bench, the three Qwen readers on StreamingBench and StreamBench, and Qwen3.8-27B on the offline benchmarks.
Our primary comparisons use training-free methods with the same frozen reader, including the training-free variant of StreamScout~\cite{streamscout}.
The tables combine published results with our reproductions, including transfers of FluxMem~\cite{fluxmem} to additional readers and SimpleStream~\cite{simplestreaming} configurations not reported in its paper.

\smallskip
\noindent\textbf{Implementation details.}
We run all experiments on 16 NVIDIA RTX 6000 Ada GPUs (48\,GB each).
\method uses a frozen V-JEPA~2 ViT-L~\cite{vjepa2} for surprisal and PE-Core-L14-336~\cite{pecore} for retrieval.
For \method, 1\,fps refers to reader-frame sampling, while V-JEPA~2 uses 4\,fps observations for token scoring.
The default configuration uses $\rho=0.3$ and $T=20$\,s for retention and $k=12$, $\tau=2.2$, and $M=4$ for the Evidence Gate.
Appendix~\ref{sec:supp_impl} gives the benchmark protocols, the remaining settings, and the prompts.

\begin{figure*}[t]
  \centering
  \includegraphics[width=\textwidth]{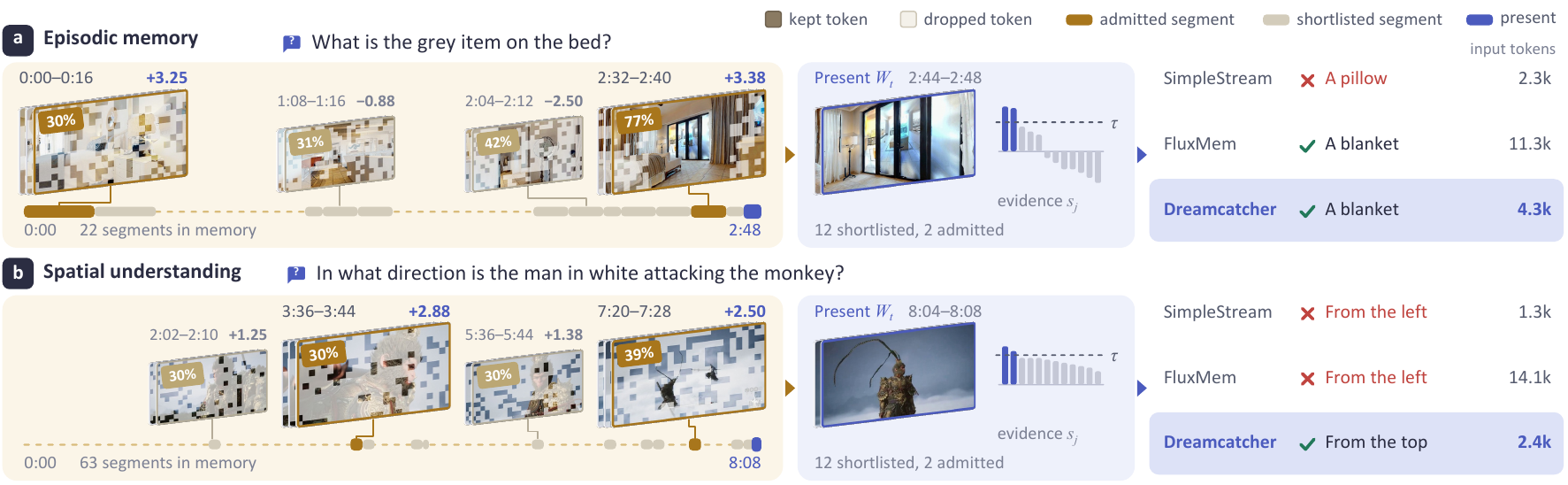}
  \vspace{-18pt}
  \caption{\textbf{Qualitative examples on OVO-Bench with Qwen3.8-27B.}
Each row shows the memory timeline for one question, with shortlisted segments marked and admitted ones in gold.
Four candidates illustrate kept and dropped tokens, time spans, retention ratios, and evidence scores $s_j$.
Beside the anchor $W_t$, bars compare all shortlisted scores with $\tau$ (dashed line).
Answers and answer-generation input tokens are shown for SimpleStream, FluxMem, and \method; counts exclude evidence scoring.}
  \label{fig:qualitative_examples}
  \vspace{-10pt}
\end{figure*}

\subsection{Benchmark Results}
\label{sec:main_results}

\smallskip
\noindent\textbf{Results on OVO-Bench.}
\method achieves the highest Overall score among training-free methods with every reader (\cref{tab:ovo_comparison}).
It exceeds the strongest training-free method in each block by 2.1--6.9 points and the corresponding 1\,fps base models by 5.7--20.9 points across the three Qwen readers.
Against FluxMem~\cite{fluxmem}, the closest token-retention method, the gains range from 3.4 to 16.2 points.
With Qwen3-VL-8B and Qwen3.8-27B, \method reaches 76.4 and 78.9, above every proprietary model in the table.

The two task groups show where these gains come from.
Reading only recent frames favors real-time perception, whereas reading the full history favors backward tracing.
With Qwen3.8-27B, SimpleStream~\cite{simplestreaming} scores 80.6 and 71.7 on the two groups, and the 1\,fps base model 70.5 and 75.9.
\method scores 79.8 and 78.0, nearly matching SimpleStream on real-time perception while achieving the best backward-tracing score, and it improves over its base model in both groups with every reader.

\smallskip
\noindent\textbf{Results on StreamingBench and StreamBench.}
\method again achieves the best overall score with all three Qwen readers (\cref{tab:comparison_stream}), by 3.6--6.7 points on StreamingBench and 1.7--7.1 points on StreamBench over the strongest baseline in each block.
With Qwen3-VL-8B and Qwen3.8-27B, it leads all four StreamingBench categories.
On the open-ended StreamBench, which shows the value of memory most clearly, the recent-window baseline falls 4.3--12.2 points below its base model, whereas \method gains 5.8--11.0 points and reaches 70.3 on long-term memory search with Qwen3.8-27B, against 31.0 for SimpleStream, so the same memory supports free-form answers as well as multiple-choice questions. \looseness=-1

\begin{table}[t]
  \centering
  \caption{\textbf{Performance comparison on offline long-video benchmarks with Qwen3.8-27B.}
    All methods sample the video at 1\,fps.
    Short, Medium, and Long are the Video-MME duration groups.
    The best score is in \textbf{bold} and the second best is \underline{underlined}.}
  \label{tab:long_video}
  \vspace{-8pt}
  \footnotesize
  \compacttable
  \fillwidth{7}{%
  \begin{tabular}{l cccc cc}
    \toprule
    \multirow{2}{*}{\textbf{Model}} & \multicolumn{4}{c}{\textbf{Video-MME}} & \multirow{2}{*}{\textbf{MLVU}} & \multirow{2}{*}{\textbf{LongVideoBench}} \\
    \cmidrule(lr){2-5}
    & Short & Medium & Long & Overall & & \\
    \midrule
    \baserow \baseicon{qwen} Qwen3.8-27B & 79.0 & 75.3 & \second{67.8} & 74.0 & 78.8 & \best{67.4} \\
    \catnone FluxMem~\cite{fluxmem} & \best{83.0} & \second{76.8} & 67.4 & \second{75.7} & \second{82.1} & 66.9 \\
    \oursrow \catnone \textbf{\method{} (Ours)} & \second{81.8} & \best{77.9} & \best{71.0} & \best{76.9} & \best{83.2} & \second{67.1} \\
    \bottomrule
  \end{tabular}}
  \vspace{-10pt}
\end{table}


\begin{figure}[t]
  \centering
  \includegraphics[width=\columnwidth]{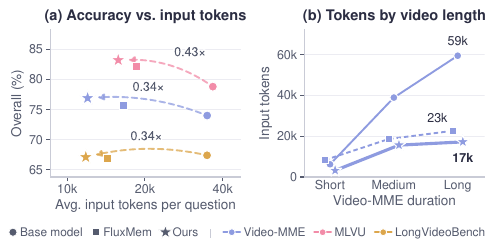}
  \vspace{-16pt}
  \caption{\textbf{Accuracy and answer-generation input tokens on offline long-video benchmarks with Qwen3.8-27B.}
    (a) Colors denote benchmarks and markers denote methods.
    Arrows run from the 1\,fps base model to \method and are labeled with \method's token ratio.
    (b) Mean input tokens per question by Video-MME duration group.}
  \label{fig:long_video_tokens}
  \vspace{-15pt}
\end{figure}

\smallskip
\noindent\textbf{Results on offline long-video benchmarks.}
We replay offline videos as streams and withhold each question until the end (\cref{tab:long_video}).
With Qwen3.8-27B, \method scores 76.9 on Video-MME and 83.2 on MLVU, improving over the 1\,fps base model by 2.9 and 4.4 points and over FluxMem by 1.2 and 1.1 points.
On LongVideoBench, it stays within 0.3 points of the base model while using 34\% of its input tokens (\cref{fig:long_video_tokens}a).
On long Video-MME videos, it scores 71.0, against 67.8 for the base model and 67.4 for FluxMem, with 17k instead of 59k input tokens (\cref{fig:long_video_tokens}b).

\begin{figure}[t]
  \centering
  \includegraphics[width=\columnwidth]{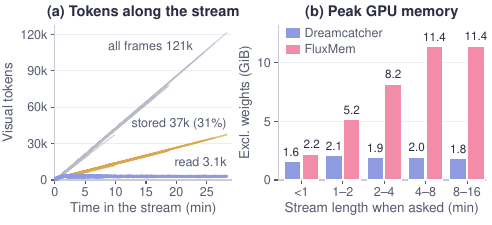}
  \vspace{-18pt}
  \caption{\textbf{Memory usage as OVO-Bench streams grow, with Qwen3.8-27B.}
(a) Visual tokens per stream: \emph{all frames} counts cumulative encoded tokens, \emph{stored} those retained in memory, and \emph{read} those of the anchor and all shortlisted segments per question.
Endpoint labels refer to the longest stream.
(b) Mean peak GPU memory by stream length at query time, excluding model weights and \method's host memory bank.}
  \label{fig:memory_vs_length}
  \vspace{-12pt}
\end{figure}

\smallskip
\looseness=-1
\noindent\textbf{Qualitative results.}
\Cref{fig:qualitative_examples} shows two questions requiring history beyond the current view.
In both, \method admits two of twelve shortlisted segments and answers correctly with far fewer input tokens than FluxMem, while SimpleStream answers incorrectly.

\subsection{Efficiency}
\label{sec:analysis}
\method reduces average answer-generation input tokens by more than 80\% relative to the 1\,fps base models of the three Qwen readers on OVO-Bench (\cref{fig:teaser}d) and uses 34--43\% of the base model's input tokens on the offline benchmarks (\cref{fig:long_video_tokens}).
As OVO-Bench streams grow (\cref{fig:memory_vs_length}), its mean peak GPU memory excluding model weights stays between 1.6 and 2.1\,GiB, whereas FluxMem's rises from 2.2 to 11.4\,GiB because its released evaluation re-encodes the latest 256 frames for each question.
\method stores each retained token in 10\,KiB of host memory, about one sixth of the 64\,KiB needed to cache the token's keys and values in the reader.
Including V-JEPA~2 and evidence scoring, the full pipeline runs 6.5 times faster than real time on two GPUs.
More details are concluded in Appendix~\ref{sec:supp_results}.

\begin{figure}[t]
  \centering
  \includegraphics[width=\columnwidth]{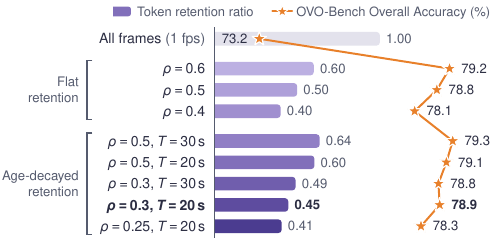}
  \vspace{-14pt}
  \caption{\textbf{Accuracy and token retention under different schedules.}
    Bars show the retained-history token ratio relative to the full 1\,fps history, and stars show OVO-Bench Overall with Qwen3.8-27B.
    Flat schedules keep a constant fraction $\rho$, and age-decayed schedules follow \cref{eq:retention}.
    Bold marks the default.}
  \label{fig:allocator_sweep}
  \vspace{-8pt}
\end{figure}

\begin{table}[t]
  \centering
  \caption{\textbf{Token selection signals at different retention floors.}
    OVO-Bench Overall with Qwen3-VL-8B and $T=20$\,s.
    Rows add random, frame-difference, or world-model surprisal scores to coverage-based selection.
    The best score in each column is in \textbf{bold} and the second best is \underline{underlined}.}
  \label{tab:token_selection}
  \vspace{-8pt}
  \footnotesize
  \compacttable
  \fillwidth{4}{%
  \begin{tabular}{l ccc}
    \toprule
    \multirow{2}{*}{\textbf{Selection signal}} & \multicolumn{3}{c}{\textbf{Retention floor} $\rho$} \\
    \cmidrule(lr){2-4}
    & 0.15 & 0.30 & 0.50 \\
    \midrule
    Coverage only & 68.0 & 68.4 & 68.8 \\
    + random & 67.9 & 68.3 & 68.9 \\
    + frame difference & \second{70.8} & \second{72.0} & \second{72.4} \\
    \oursrow \textbf{+ world-model surprisal} & \best{74.7} & \best{76.4} & \best{76.9} \\
    \bottomrule
  \end{tabular}}
  \vspace{-8pt}
\end{table}

\subsection{Ablations and Analysis}
\label{sec:ablation}
\looseness=-1
We ablate each component of \method on OVO-Bench, changing one at a time and keeping the others at their default settings.
We compare selection signals with two readers, Qwen3-VL-8B and Qwen3.8-27B, and analyze the retention schedule and the Evidence Gate with Qwen3.8-27B.

\smallskip
\looseness=-1
\noindent\textbf{World-model surprisal versus simpler signals.}
World-model surprisal consistently outperforms the alternative selection signals across the three retention floors with Qwen3-VL-8B (\cref{tab:token_selection}), improving over frame differences by 3.9--4.5 points and over coverage alone by 6.7--8.1 points.
At $\rho=0.15$, it reaches 74.7, exceeding frame differences at $\rho=0.50$ by 2.3 points.
Qwen3.8-27B also shows consistent improvements, with smaller gains reported in Appendix~\ref{sec:supp_ablation_detail}.

\smallskip
\noindent\textbf{Retention schedule.}
\Cref{fig:allocator_sweep} compares flat retention with budgets that decay with segment age.
All eight schedules score between 78.1 and 79.3 while keeping 40--64\% of the history tokens, above the 73.2 of reading all frames at 1\,fps, so \method is robust to the exact schedule.
Age decay uses these tokens efficiently, as the default keeps 45\% of them and scores 78.9, matching flat 50\% retention (78.8).
We use this default because it balances accuracy against retained tokens rather than maximizing accuracy.
The most accurate schedule, $\rho=0.5$ with $T=30$\,s, scores 0.4 points higher but keeps 64\% of the tokens.

\begin{figure}[t]
  \centering
  \includegraphics[width=\columnwidth]{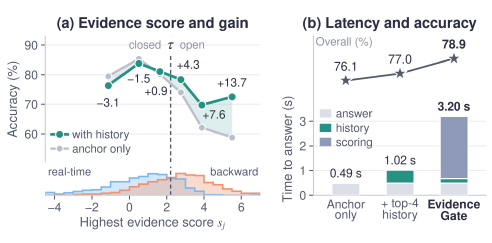}
  \vspace{-18pt}
    \caption{\textbf{Evidence Gate analysis on OVO-Bench with Qwen3.8-27B.}
      (a) Accuracy from the anchor alone and with history, for questions binned by their highest evidence score.
      Labels give the gain from history, the dashed line marks $\tau$, and histograms show the score distribution of each task group.
      (b) Mean time to answer by stage, with Overall accuracy as stars.
      Top-4 reads the four highest-ranked segments without evidence scoring.}
  \label{fig:gate_ablation}
  \vspace{-10pt}
\end{figure}

\smallskip
\noindent\textbf{Evidence scores track the value of history.}
We bin the questions by their highest shortlisted evidence score $s_{\max}$ and compare answering from the anchor alone with reading history (\cref{fig:gate_ablation}a), using diagnostic runs that admit more candidates below the default threshold.
The benefit of history rises steadily with the score, from $-3.1$ points in the lowest bin to $+13.7$ in the highest, so low-scoring history hurts even after similarity retrieval, which motivates scoring each candidate against the current view.
The gate receives no task labels, yet backward-tracing questions tend to score higher.

\smallskip
\noindent\textbf{Accuracy and cost of the gate.}
On all 1,468 questions (\cref{fig:gate_ablation}b), answering from the anchor alone scores 76.1, reading the top four retrieved segments without scoring reaches 77.0, and the Evidence Gate reaches 78.9, about three times the gain of unscored retrieval.
Mean latency rises from 0.49\,s with the anchor alone and 1.02\,s with unscored top-4 history to 3.20\,s with the gate, mostly from about ten scoring passes per question, which stay short because they reuse the anchor's cache and generate no tokens.

\section{Conclusion}
\label{sec:conclusion}

We introduced \method, a training-free framework in which the surprisal of an independently pretrained, frozen video world model guides which native visual tokens are retained, and an Evidence Gate admits only history that adds evidence beyond the current view.
\method achieves the highest overall scores among the compared training-free methods with every evaluated Video-LLM on three streaming and three offline benchmarks, reaching 78.9 on OVO-Bench, 79.7 on StreamingBench, and 75.1 on StreamBench with Qwen3.8-27B.
On OVO-Bench, it also reduces answer-generation input tokens by more than 80\% relative to the 1\,fps Qwen base models.
Ablations show that world-model surprisal improves over frame differences by 3.9--4.5 points (Qwen3-VL-8B) and that evidence scoring improves over unscored retrieval by 1.9 points (Qwen3.8-27B).


\section*{LLM Usage Statement}
The authors conceived the ideas, designed the method and all experiments, and take full responsibility for the paper and its results.
We used large language model assistants (ChatGPT and Claude) to polish the language and to draft and revise text under the authors' direction, and the authors reviewed and edited all of it.
Claude also assisted in implementing and running the experiments, and analyzing the results.
The authors verified every cited reference against its original source.
\section*{Acknowledgments}
This work has been funded in part by the ARL award W911QX-24-F-0049, DARPA award FA8750-23-2-1015, ONR award N00014-23-1-2840, ONR MURI grant N00014-25-1-2116, and USDA-NIFA award 2021-67021-35329.
\par
{
  \small
  \bibliographystyle{ieeenat_fullname}
  \bibliography{main}
}
\clearpage
\iftoggle{cvprfinal}{}{%
  \hypersetup{pageanchor=false}%
  \backrefsetup{disable}%
  \setcounter{page}{1}%
}
\appendix
\raggedbottom

\renewcommand{\thefigure}{\Alph{section}\arabic{figure}}
\renewcommand{\thetable}{\Alph{section}\arabic{table}}
\renewcommand{\theHfigure}{appendix.\Alph{section}.\arabic{figure}}
\renewcommand{\theHtable}{appendix.\Alph{section}.\arabic{table}}
\counterwithin*{figure}{section}
\counterwithin*{table}{section}
\setcounter{figure}{0}
\setcounter{table}{0}

\maketitlesupplementary

\section*{Overview}
\begingroup
\setlength{\parskip}{0pt}
We provide additional experiments on \method, explain the design of its streaming memory, and document how we implement and evaluate it.

\begin{itemize}[leftmargin=*,topsep=3pt,itemsep=4pt,partopsep=0pt]
  \item \textbf{Section~\ref{sec:supp_results}. More Experimental Results.}
  We break down performance by benchmark and task and examine how token selection, the retention schedule, and the Evidence Gate affect accuracy.
  In Section~\ref{sec:supp_efficiency}, we report the cost of forming memory and answering questions, including token counts, latency, and memory usage.
  We close with more qualitative examples on OVO-Bench that show which earlier segments the gate admits and how the reader answers individual questions.

  \item \textbf{Section~\ref{sec:supp_discussions}. Further Discussion of the Method.}
  We derive how the retention budget grows with stream duration and explain what bounds the amount of history the reader inspects per question.
  We describe how V-JEPA~2 predicts incoming video in feature space and measure how quickly the system turns each new block into visual memory.
  We then discuss how the system retains visual details before a question is known and selects evidence from that memory once the question arrives.

  \item \textbf{Section~\ref{sec:supp_impl}. More Implementation Details.}
  We define the benchmark tasks and scoring rules in Section~\ref{sec:supp_benchmarks}.
  We then describe how we construct the visual anchor, compute surprisal, allocate tokens as segments age, and score candidate segments with each frozen reader.
  In Section~\ref{sec:supp_prompts}, we show how we arrange the visual evidence and dialogue history in the prompt and how we score multiple-choice and open-ended answers.

  \item \textbf{Section~\ref{sec:supp_impact}. Limitations and Broader Impact.}
  We describe the GPU memory and reader interfaces required for deployment, and discuss the uses and privacy implications of query-agnostic memory for long-running live streams.
\end{itemize}
\endgroup

\section{More Experimental Results}
\label{sec:supp_results}


\subsection{Benchmark Results in Detail}
\label{sec:supp_bench_detail}
Appendix~\ref{sec:supp_benchmarks} defines the streaming tasks and explains how each benchmark aggregates their scores.

\smallskip
\noindent\textbf{OVO-Bench.}
\Cref{tab:ovo_comparison} compares \method with training-free methods that use the same reader.
Its Overall margin over the strongest training-free method in each block is 3.2 points with LLaVA-OneVision-7B and 2.1 points with Qwen2.5-VL-7B, both over SimpleStream, 6.9 points with Qwen3-VL-8B over ShallowStream, and 2.8 points with Qwen3.8-27B over SimpleStream.
Over its own base model, the gains are 8.2, 9.3, 20.9, and 5.7 points, and over FluxMem they are 4.6, 4.2, 16.2, and 3.4 points.
With LLaVA-OneVision, whose visual tokenization differs from that of the Qwen readers, \method leads all six real-time tasks and reaches a real-time average of 73.1, 3.5 points above the next method.
With Qwen2.5-VL-7B, it leads the real-time average with 75.7, 4.3 points above SimpleStream.
With Qwen3-VL-8B, its backward-tracing average of 71.9 exceeds the next best method, StreamScout, by 11.8 points, and it ranks second on both episodic memory (61.3) and action sequence identification (71.6).
It also leads spatial understanding with 72.5.
With Qwen3.8-27B, it achieves the best episodic-memory score of 72.1, 6.8 points above the base model and 7.4 points above FluxMem.
It also leads OCR with 95.3 and ranks second on HLD with 95.7, below SimpleStream's 96.8.

\smallskip
\noindent\textbf{OVO-Bench excluding HLD.}
\Cref{tab:ovo_no_hld} gives equal weight to the real-time average and the mean of EPM and ASI, using the rounded task scores in \cref{tab:ovo_comparison}.
With HLD excluded, \method still leads the compared training-free methods for each of the four readers.

\smallskip
\noindent\textbf{StreamingBench.}
On StreamingBench (\cref{tab:comparison_stream}), \method improves the overall score over the strongest baseline in each block by 3.6 points with Qwen2.5-VL-7B over StreamScout, 5.9 points with Qwen3-VL-8B over SimpleStream, and 6.7 points with Qwen3.8-27B over FluxMem.
Over the base models, the gains are 10.7, 10.5, and 8.7 points.
With Qwen3-VL-8B and Qwen3.8-27B, it leads all four reported categories.
With Qwen3.8-27B, it improves over FluxMem by 5.9 points on real-time visual understanding, 8.8 points on anomaly context understanding, 12.0 points on misleading context understanding, and 6.8 points on sequential question answering.

\begin{table}[t]
  \centering
  \caption{\textbf{OVO-Bench excluding HLD.}
    Approximate Overall (\%) recomputed from the rounded scores in \cref{tab:ovo_comparison}, with equal weights for the real-time average and the mean of EPM and ASI.
    Each row compares against the strongest training-free baseline under this metric.
    The best score in each row is in \textbf{bold}.}
  \label{tab:ovo_no_hld}
  \vspace{-6pt}
  \footnotesize
  \compacttable
  \fillwidth{3}{%
  \begin{tabular}{l lc}
    \toprule
    \textbf{Reader} & \textbf{Strongest baseline} & \textbf{Ours} \\
    \midrule
    \baseicon{llava} LLaVA-OneVision-7B & HERMES (63.7) & \cellcolor{oursbg}\best{63.9} \\
    \baseicon{qwen} Qwen2.5-VL-7B & QueryStream (62.2) & \cellcolor{oursbg}\best{64.1} \\
    \baseicon{qwen} Qwen3-VL-8B & ShallowStream (71.7) & \cellcolor{oursbg}\best{73.7} \\
    \baseicon{qwen} Qwen3.8-27B & FluxMem (73.3) & \cellcolor{oursbg}\best{74.5} \\
    \bottomrule
  \end{tabular}}
\end{table}

\smallskip
\noindent\textbf{StreamBench.}
The benefit of retained history is particularly clear on long-term memory search, where the question arrives more than a minute after the event ends (\cref{sec:supp_benchmarks}).
With Qwen3.8-27B, \method scores 70.3 on this task, compared with 58.1 for the 1\,fps base model, 54.8 for FluxMem, and 31.0 for the four-frame SimpleStream baseline (\cref{tab:comparison_stream}). With Qwen3-VL-8B, \method scores 55.8 compared with 24.1 for SimpleStream.
In the benchmark average, \method leads the strongest baseline by 1.7 points with Qwen2.5-VL-7B over FluxMem, 1.7 with Qwen3-VL-8B over StreamScout, and 7.1 with Qwen3.8-27B over FluxMem. Its gains over the corresponding base models are 5.8, 8.3, and 11.0 points, while SimpleStream falls below the same base models by 4.3, 12.2, and 9.3 points.
With Qwen3.8-27B, \method leads five of the six categories, including object search (66.6 vs.\ 53.8 for FluxMem), short-term memory search (69.6 vs.\ 64.8), and conversational interaction (85.2 vs.\ 77.7).

\smallskip
\noindent\textbf{Offline long-video benchmarks.}
On Video-MME with Qwen3.8-27B, longer videos yield greater savings in answer-generation input tokens relative to the 1\,fps base model (\cref{fig:long_video_tokens}b).
On short videos, \method uses about 52\% of the base model's input (3.2k of 6.2k tokens per question). On long videos, it uses about 29\% (17.2k of 59.4k).
Accuracy improves over the same base model in all three duration groups, by 2.8 points on short videos, 2.6 on medium videos, and 3.2 on long videos (\cref{tab:long_video}).
On long videos, \method scores 71.0, compared with 67.8 for the base model and 67.4 for FluxMem.
On MLVU, it improves accuracy over the base model by 4.4 points and over FluxMem by 1.1 points.

\subsection{Ablations in Detail}
\label{sec:supp_ablation_detail}

\smallskip
\noindent\textbf{Token selection.}
World-model surprisal gives the highest OVO-Bench Overall among the tested selection signals at all three retention floors with both Qwen3-VL-8B and Qwen3.8-27B (\cref{tab:token_selection,tab:token_selection_qwen38}).
With Qwen3.8-27B, surprisal at a 15\% floor already matches frame differences at a 50\% floor (77.8 Overall).
At matched floors of 15\%, 30\%, and 50\%, its gains over frame differences are 0.9, 1.8, and 1.3 points, and its gains over coverage alone are 1.2, 2.0, and 2.0 points.
Frame differences score each token by its change from the previous frame. World-model surprisal scores the normalized difference between the features predicted from preceding observations and those observed in the current block (\cref{eq:residual}).
In the surprisal maps in \cref{fig:teaser}a, the crossing and lifting hands receive higher scores than most of the background.
The coverage share also samples across the full token grid, preserving predictable details that a later question may need.

\begin{table}[!htbp]
  \centering
  \caption{\textbf{Token selection signals at different retention floors.}
    OVO-Bench Overall with Qwen3.8-27B and $T=20$\,s.
    Rows add random, frame-difference, or world-model surprisal scores to coverage-based selection.
    The best score in each column is in \textbf{bold} and the second best is \underline{underlined}.}
  \label{tab:token_selection_qwen38}
  \vspace{-8pt}
  \footnotesize
  \compacttable
  \fillwidth{4}{%
  \begin{tabular}{l ccc}
    \toprule
    \multirow{2}{*}{\textbf{Selection signal}} & \multicolumn{3}{c}{\textbf{Retention floor} $\rho$} \\
    \cmidrule(lr){2-4}
    & 0.15 & 0.30 & 0.50 \\
    \midrule
    Coverage only & 76.6 & 76.9 & 77.1 \\
    + random & 76.4 & 76.6 & 77.2 \\
    + frame difference & \second{76.9} & \second{77.1} & \second{77.8} \\
    \oursrow \textbf{+ world-model surprisal} & \best{77.8} & \best{78.9} & \best{79.1} \\
    \bottomrule
  \end{tabular}}
\end{table}

\smallskip
\noindent\textbf{Retention schedule.}
We choose the default schedule ($\rho=0.3$, $T=20$\,s) to balance accuracy and compression of the visual history (\cref{fig:allocator_sweep}).
It scores 78.9 on OVO-Bench while retaining 45\% of the history tokens at question time. The highest-scoring schedule tested ($\rho=0.5$, $T=30$\,s) retains 64\% of the history, an additional 19 percentage points, for an accuracy gain of only 0.4 points (79.3 Overall).
Across all eight schedules, Overall ranges from 78.1 to 79.3 at mean retention ratios of 40--64\%, compared with 73.2 for reading all frames at 1\,fps.
Flat and age-decayed schedules give similar scores at comparable retention. Flat 50\% retention scores 78.8, close to the default's 78.9. At a tighter budget, the decaying schedule with $\rho=0.25$ and $T=20$\,s retains 41\% and scores 78.3, compared with 78.1 for flat 40\% retention.
For age-decayed schedules, recent segments retain more detail, so the mean retention ratio over questions exceeds the floor.

\smallskip
\noindent\textbf{Evidence scores.}
\Cref{fig:gate_ablation}a groups the OVO-Bench questions into six bins of their highest shortlisted evidence score, with cutoffs at 0, 1.1, 2.2, 3.5, and 5.0, and reports accuracy within each bin.
For the four bins at and above 1.1, the history results come from a diagnostic run with $\tau=1.1$, so they measure the value of reading history rather than the output of the default gate.
The benefit of history grows steadily across the six bins, from $-3.1$ and $-1.5$ points to $+0.9$, $+4.3$, $+7.6$, and $+13.7$ points.
In the two highest bins, the anchor alone answers only about 60\% of the questions correctly, compared with 80--85\% in the two lowest bins, so a high score marks both a greater need for evidence and a larger benefit from history.
Just below the default threshold, reading history adds only 0.9 points, so the default gate stays closed there and saves the history processing.
The gate receives no task labels.
Backward-tracing questions tend to receive higher scores than real-time ones, but the two distributions overlap, so some real-time questions read history and some backward-tracing questions use the anchor alone.
Admission therefore follows the evidence available for each question.

\smallskip
\noindent\textbf{Gate accuracy and cost.}
On all 1,468 questions, the anchor alone scores 76.1 Overall, reading the top four retrieved segments without scoring scores 77.0, and the Evidence Gate scores 78.9 (\cref{fig:gate_ablation}b).
Retrieval alone adds 0.9 points, and evidence-based admission adds a further 1.9 points.
The gate selectively admits history to the final answering context, up to the cap of $M=4$.
This selectivity reduces the time spent processing admitted history relative to top-4 reading.
Evidence scoring adds about ten passes per question that share the anchor as a common prefix, and \cref{sec:supp_efficiency} details the timing.

\subsection{Efficiency Measurement Details}
\label{sec:supp_efficiency}
We separate the context used to generate an answer from the cost of scoring candidate history and maintaining the memory bank, since these quantities measure different parts of the system and need not decrease in proportion.

\smallskip
\noindent\textbf{Answer-generation input tokens.}
Token counts cover the input of the final answering pass and are averaged over questions.
They exclude the evidence-scoring passes, which are reported separately as part of the query latency.
The offline token ratios in \cref{fig:long_video_tokens} use the base-model input defined in Appendix~\ref{sec:supp_benchmarks}.
On Video-MME, the input of \method grows from 3.2k tokens on short videos to 17.2k on long videos, while the base model grows from 6.2k to 59.4k and FluxMem uses 22.6k on long videos.

\begin{figure}[t]
  \centering
  \includegraphics[width=\linewidth]{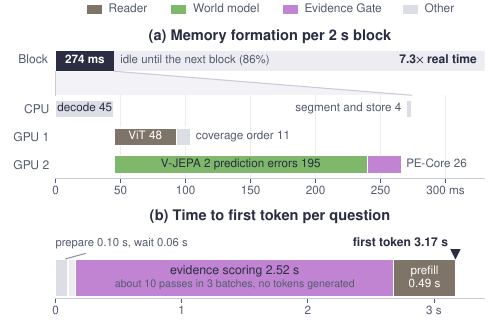}
  \caption{\textbf{Runtime profile of \method.}
    Qwen3.8-27B on two RTX 6000 Ada GPUs, with times averaged over OVO-Bench.
    (a) Memory formation for one 2\,s block, in ms.
    The top bar spans the block, and the lanes below expand its busy part by device.
    (b) Time to first token for one question, split by stage.}
  \label{fig:runtime}
\end{figure}

\smallskip
\noindent\textbf{Query latency.}
The latencies in \cref{fig:gate_ablation}b are means over all 1,468 OVO-Bench questions with Qwen3.8-27B.
They measure the time to complete an answer and include evidence scoring whenever it is used.
Mean total latency is 0.49\,s for the anchor alone, 1.02\,s for top-4 history without evidence scoring, and 3.20\,s for the Evidence Gate.
The breakdown separates answer generation, the processing of admitted history, and evidence scoring.
The Evidence Gate reduces history processing by admitting history for fewer questions and fewer segments per question.
Evidence scoring accounts for most of its query time, with about ten scoring passes per question sharing the anchor as a common prefix.
The Qwen3.8-27B value in \cref{fig:teaser}d is the mean time to first token in \cref{fig:runtime}, 3.17\,s, measured with the streaming implementation, which scores candidates in batches of up to four (Appendix~\ref{sec:supp_gate}).
The OASIS~\cite{oasis} and StreamScout~\cite{streamscout} timings in that panel are taken from their papers and were measured on an A800 and an A100, so the panel places the methods in a common deployment context rather than on matched hardware.

\smallskip
\noindent\textbf{Model placement and weights.}
Each Qwen3.8-27B run uses two RTX 6000 Ada GPUs.
The reader is split by layer, with its vision encoder, token embeddings, and the first 29 of its 64 decoder layers on the first GPU and the remaining layers and the output head on the second.
V-JEPA~2 and PE-Core run on the second GPU, where a worker thread encodes each block in parallel with the reader's vision encoder on the first GPU.
After loading and before any video is processed, the bf16 weights of the reader occupy 50.96\,GiB, split into 23.79\,GiB and 27.17\,GiB across the two GPUs.
V-JEPA~2 ViT-L, with its context encoder, predictor, and target encoder in bf16, occupies 1.17\,GiB, and PE-Core-L14-336 in fp32 occupies 2.50\,GiB, so all models together occupy 54.64\,GiB.
These weights are a fixed cost that does not grow with the stream.
FluxMem uses the same reader with the same split, so before any stream is processed the two methods differ by the 3.7\,GiB of encoder weights.

\smallskip
\noindent\textbf{Runtime profile.}
\Cref{fig:runtime} profiles \method with Qwen3.8-27B on OVO-Bench, with times averaged over all questions and their streams.
Each 2\,s block is decoded once and then encoded on both GPUs in parallel.
The first GPU runs the reader's vision encoder and computes the coverage order, while V-JEPA~2 and PE-Core process the same block on the second GPU.
After segmentation and storage, a block takes 274\,ms, or 14\% of its duration, so memory formation runs 7.3 times faster than real time.
V-JEPA~2 accounts for 195\,ms of this time.
Answers are computed on a separate thread from the anchor and the shortlist gathered when the question arrives, so memory formation continues while a question is answered.
The mean time to first token is 3.17\,s, measured from the moment the reader receives the question.
Evidence scoring takes 2.52\,s of it, with about ten passes in three batches that generate no tokens, and answer prefill takes 0.49\,s.
The remaining 0.16\,s covers preparing the anchor and the shortlist and waiting for an earlier question of the same stream.
OVO-Bench poses about 37 questions per hour of video, so answering them adds about 3\% of real time to the 14\% spent on memory formation.
An accelerated replay of 94 streams, which feeds each block as soon as the previous one is done and answers questions while memory formation continues, runs 6.5 times faster than real time.

\smallskip
\noindent\textbf{Memory profile.}
\Cref{fig:memory_vs_length} profiles the full OVO-Bench evaluation with Qwen3.8-27B.
In panel (a), the curves count all visual tokens of the encoded frames and the tokens retained in the memory bank, and the dots count the tokens that the reader inspects for a question.
This read count includes the anchor and every shortlisted segment evaluated by the Evidence Gate, so it is larger than the admitted context used in the final answering pass.
At the end of the plotted stream, the bank stores about 37k of the 121k encoded visual tokens, while the reader inspects about 3.1k tokens for the question.
The bank thus preserves a growing record, while retrieval bounds how much of it the reader inspects.
Panel (b) reports peak GPU memory from the start of each stream, covering both memory formation and answering, and excludes the fixed weights of the reader, V-JEPA~2, and the retrieval encoder.
FluxMem's released evaluation re-encodes its latest 256 frames for each question, so its peak usage rises until this window is full.
Including model weights, total peak GPU memory after four minutes is 56.6\,GiB for \method and 62.4\,GiB for FluxMem, despite the 3.7\,GiB of additional encoder weights in \method.
These totals are summed over the two GPUs that host each method.
The memory bank of \method resides in host memory, where each retained token occupies about 10\,KiB, or about 14\,MiB per minute of stream in this profile.
Each retained token is a 5,120-dimensional bf16 embedding at the input of the reader's language model.
Caching its keys and values instead would take 64\,KiB, since Qwen3.8-27B keeps them in 16 full-attention layers with four 256-dimensional heads each, while its other 48 layers use Gated DeltaNet states that do not grow with the number of tokens.
The bank therefore needs about one sixth of the storage of a key--value cache over the same tokens.

\subsection{More Qualitative Examples}
\label{sec:supp_qualitative}

\begin{figure*}[t]
  \centering
  \includegraphics[width=\textwidth]{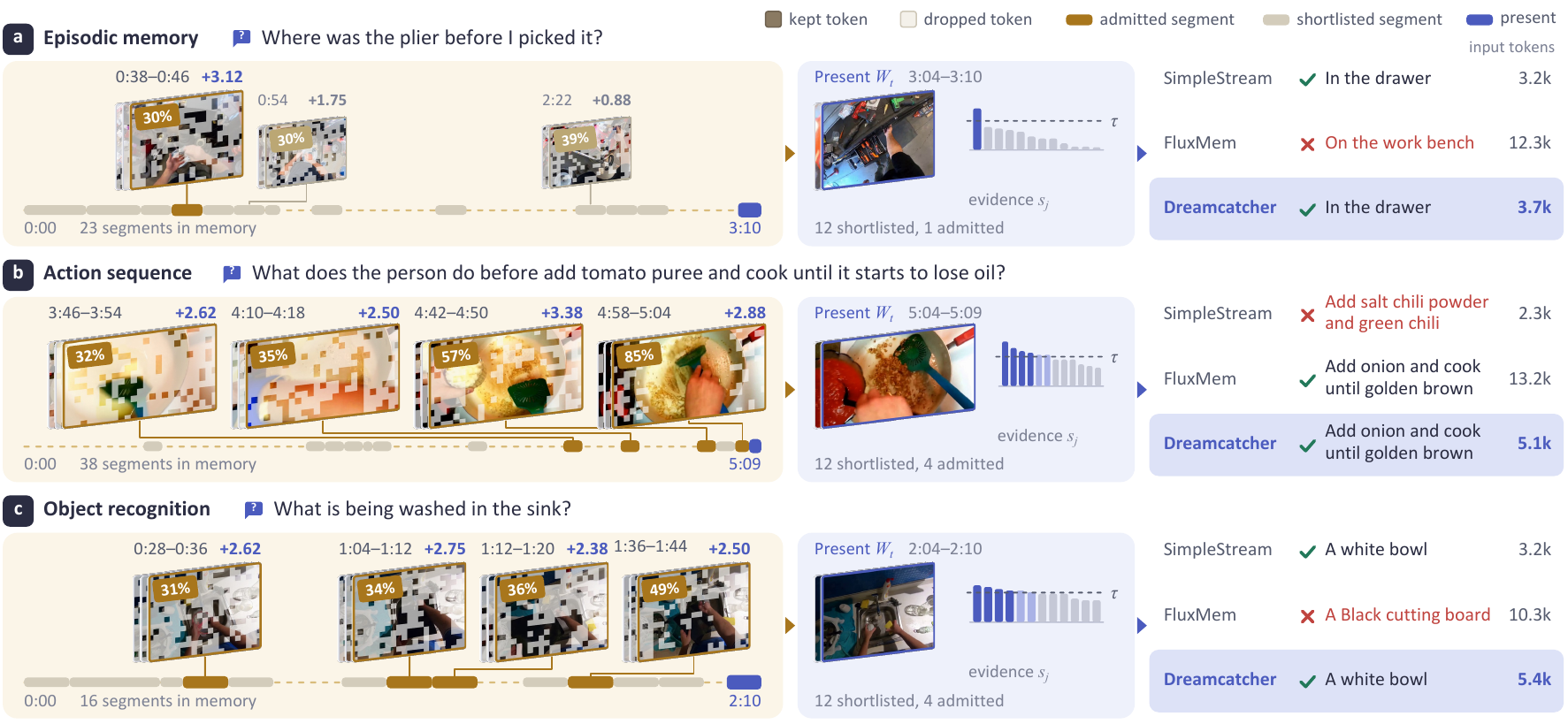}
  \caption{\textbf{More qualitative examples on OVO-Bench with Qwen3.8-27B.}
    The rows show (a) episodic memory, (b) action sequence identification, and (c) object recognition.
    The percentages give the share of visual tokens retained in each labeled segment at question time. The scores $s_j$ show how the reader rates each candidate's evidence given the question and present anchor $W_t$.
    Gold marks admitted segments, and the dashed line marks the admission threshold $\tau$. Input-token counts cover the final answering pass.}
  \label{fig:qualitative_examples_more}
\end{figure*}

\noindent\textbf{Surprisal guides which visual details enter memory.}
In \cref{fig:teaser}a, the hands receive higher prediction errors than most of the background when they cross at 0:46 and when one hand lifts at 1:56.
The allocator uses these spatial differences to give poorly predicted patches priority within the surprisal share of a segment's token budget, before any question about them is known. The ablation in \cref{tab:token_selection_qwen38} measures the value of this signal at the default retention floor, where surprisal scores 78.9 Overall and frame differences score 77.1 with the same reader.
Once a question arrives, the Evidence Gate scores how much each shortlisted segment adds to the current view. \Cref{fig:qualitative_examples_more} shows these question-time decisions for three additional OVO-Bench questions with Qwen3.8-27B.

\smallskip
\noindent\textbf{Reading several moments to recover an earlier step.}
At 5:09, the question in \cref{fig:qualitative_examples_more}b asks what the cook did before adding tomato puree. The anchor covers 5:04--5:09, and SimpleStream answers with a later step, ``Add salt chili powder and green chili.''
From 38 stored segments, \method shortlists twelve and admits four earlier intervals between 3:46 and 5:04. The oldest retains 32\% of its tokens and the most recent retains 85\%. All four receive evidence scores above the admission threshold ($+2.50$ to $+3.38$).
Reading those intervals together with the anchor, \method answers ``Add onion and cook until golden brown'' with 5.1k input tokens. FluxMem gives the same correct answer with 13.2k.

\smallskip
\noindent\textbf{Reading history for object questions.}
When asked at 3:10 where the pliers were before they were picked up (\cref{fig:qualitative_examples_more}a), the gate admits one of twelve candidates. The 0:38--0:46 segment still holds about 30\% of its visual tokens and scores $+3.12$.
\method and SimpleStream both answer ``In the drawer,'' with 3.7k and 3.2k input tokens, respectively. FluxMem answers ``On the work bench'' with 12.3k.
For the question about the object being washed (\cref{fig:qualitative_examples_more}c), the gate admits four segments. \method answers ``A white bowl'' with 5.4k tokens. SimpleStream also identifies the bowl, and FluxMem identifies a black cutting board with 10.3k tokens.

\smallskip
\noindent\textbf{How much history reaches the answer.}
In the two main-text examples (\cref{fig:qualitative_examples}), the gate admits two segments per question. \method identifies the grey item on the bed as a blanket and the direction of attack as ``From the top.''
Across those two questions and the three in \cref{fig:qualitative_examples_more}, \method answers all five correctly with 2.4k--5.4k input tokens in the final answering pass, compared with 10.3k--14.1k for FluxMem.
The two questions in \cref{fig:teaser}c show how the same gate changes its decision within a single stream. It admits the 1:12--1:20 segment for the question about cutting dough. For the towel question, all candidates score below threshold and the reader answers ``on the table'' from the anchor alone.

\needspace{9\baselineskip}
\section{Further Discussion of the Method}
\label{sec:supp_discussions}

\subsection{Compute and Memory Cost}
\label{sec:supp_cost}

For a stream of length $L$ with a constant rate of $n$ candidate tokens per second, integrating the retention schedule in \cref{eq:retention} gives the idealized retention budget
\begin{equation}
  n \int_0^{L} r(a)\, \mathrm{d}a = n \left[ \rho L + (1 - \rho)\, T \left( 1 - e^{-L/T} \right) \right]
  \label{eq:memory_size}
\end{equation}
in tokens, ignoring segment discretization and rounding.
Dividing by $nL$ gives the mean retention ratio at the end of a stream, which is 52.2\% for a 1\,min video and 30.4\% for a 1\,h video under the default schedule.
The long-run token rate is $\rho n$; the exponential term adds at most $n(1-\rho)T$ tokens in this approximation.
Physical storage additionally includes tokens awaiting safe deletion and working buffers.

As the stream grows, the bank retains a longer visual history while the reader inspects a bounded amount of visual input per question.
Shortlisting searches one retrieval key per eligible segment. At the mean segment duration of 7.8\,s in the OVO-Bench profile, an hour of video would add about 460 keys to the index.
With the default settings, the reader scores at most $k=12$ segments against the anchor and uses at most $M=4$ of them alongside the anchor to generate its answer. Each candidate spans at most 16\,s. \Cref{sec:supp_efficiency} reports the measured costs.

\subsection{Computing Surprisal During Memory Formation}
\label{sec:supp_world_model}

Given the preceding frames, frozen V-JEPA~2 efficiently predicts each incoming video block in a learned feature space~\cite{vjepa2}.
For each spatiotemporal patch, we compare its predicted and observed features, then pool the normalized errors onto the reader's token grid.
The same frozen predictor guides all four readers, each of which retains its own native visual tokens.

On OVO-Bench with Qwen3.8-27B and two RTX 6000 Ada GPUs, the V-JEPA~2 stage takes 195\,ms on average per 2\,s video block and runs in parallel with the reader's vision encoder.
The complete memory update takes 274\,ms on average, about 14\% of the block's duration (\cref{fig:runtime}). The V-JEPA~2 weights occupy 1.2\,GiB.

\subsection{Forming Memory and Selecting Evidence}
\label{sec:supp_write_read}

Before a question is known, coverage preserves a range of visual details in every block, and surprisal directs the remaining budget toward details the world model predicted poorly. As a segment ages, its retention floor keeps a subset of those observations available for later questions.
Once a question arrives, the Evidence Gate asks whether each shortlisted segment adds evidence beyond the current view. The same bank can supply several earlier moments for a question about a sequence of actions, or the reader can answer from the anchor alone when no candidate reaches the threshold (\cref{fig:qualitative_examples_more}b and \cref{fig:teaser}c).

\section{More Implementation Details}
\label{sec:supp_impl}

\subsection{Benchmarks and Evaluation}
\label{sec:supp_benchmarks}
For the tasks evaluated below, we follow each benchmark's question format, answer format, and scoring protocol. Streaming questions are answered at their annotated times from the observed prefix. The OVO-Bench memory profile in \cref{fig:memory_vs_length}a covers a stream shorter than half an hour. The offline benchmarks extend the evaluation to longer videos through end-of-video questions.

\smallskip
\noindent\textbf{OVO-Bench.}
OVO-Bench~\cite{ovobench} poses each question at a timestamp of the video and evaluates the answer using only the frames observed up to that point.
We evaluate its real-time visual perception and backward-tracing splits, which contain 1,468 multiple-choice questions in total.
The real-time tasks ask about the present scene.
Optical character recognition (OCR) reads visible text, action recognition (ACR) identifies the ongoing action, attribute recognition (ATR) asks about properties of visible objects, spatial understanding (STU) asks about spatial relations, future prediction (FPD) asks what is likely to happen next, and object recognition (OJR) identifies objects in view.
The backward-tracing tasks depend on earlier observations.
Episodic memory (EPM) asks about past events and objects, action sequence identification (ASI) asks for the order of earlier actions, and hallucination detection (HLD) asks about content that may not have appeared in the stream.
Task accuracies are averaged within each group, and Overall is the mean of the two group averages.

\smallskip
\noindent\textbf{StreamingBench.}
StreamingBench~\cite{streamingbench} poses multiple-choice questions at specific times of a stream.
Real-time visual understanding (RTVU) aggregates its real-time tasks, which cover perception of objects, attributes, actions, and spatial relations, counting, text-rich understanding, event understanding, causal and prospective reasoning, and clip summarization.
Anomaly context understanding (ACU) asks the model to identify an unexpected event, such as a brief interruption in an otherwise repetitive activity.
Misleading context understanding (MCU) asks about the current scene when earlier frames show a similar but outdated state.
Sequential question answering (SQA) asks related questions in sequence over the same stream.
For SQA, the benchmark supplies the earlier questions, options, and annotated answers as context for the current turn.
Proactive output, which requires the model to decide when to respond, lies outside our question-answering setting.
Avg weights task scores by their question counts.

\smallskip
\noindent\textbf{StreamBench.}
StreamBench~\cite{streamchat} poses open-ended questions during a stream. The reader generates a free-form answer at each turn, and a separate judge compares it with the reference (\cref{sec:supp_prompts}).
Object search (OS) asks where an object is or was.
In long-term memory search (LM), the event lasts more than 5\,s and the question arrives more than 1\,min after the event ends. In short-term memory search (SM), the question arrives less than 20\,s after the event ends.
Conversational interaction (CI) continues a dialogue about the video, knowledge-based questions (KG) draw on world knowledge about what is shown, and simple factual questions (SF) ask about directly observable facts.
Avg weights task scores by their question counts.

\smallskip
\noindent\textbf{Offline long-video benchmarks.}
Each video is replayed as a stream at 1\,fps, memory is formed without access to the question, and the question is posed at the end of the video.
For Video-MME~\cite{videomme}, we report accuracy without subtitles on all videos and on its official short, medium, and long duration groups.
For MLVU~\cite{mlvu}, we report M-Avg, the mean accuracy over the multiple-choice tasks of its development set.
For LongVideoBench~\cite{longvideobench}, we report accuracy on the validation set.
The 1\,fps base model reads the video uniformly subsampled to at most 768 frames.

\smallskip
\noindent\textbf{Baselines and reporting.}
We report the training-free variant of StreamScout~\cite{streamscout} throughout.
The Qwen base models read the observed video at 1\,fps, and the LLaVA-OneVision base model reads 64 frames.
SimpleStream reads the four most recent frames; \method uses the anchor defined in \cref{sec:supp_visual}.
On Qwen3.8-27B, the released drop threshold of FluxMem, 0.8, keeps fewer than 10\% of the visual tokens, so we raise it to 0.97, which brings its retention closer to that of its default setting on Qwen2.5-VL.
Accuracy differences are given in percentage points, and comparisons are made within each reader block.

\subsection{Visual Tokens and Anchor}
\label{sec:supp_visual}
The reader samples frames at 1\,fps, and memory construction uses 2\,s blocks aligned with the V-JEPA~2 prediction targets.
For the Qwen readers, each block coincides with a two-frame temporal patch of the vision encoder; for LLaVA-OneVision, the two frames are encoded individually within the same block.
The visual anchor retains the latest 4--7 sampled frames, with its start aligned to a 4\,s grid, or the available prefix at startup.
Its visual tokens are retained in full.
Incomplete Qwen temporal patches are padded by repeating the last available frame.

\subsection{World-Model Surprisal}
\label{sec:supp_surprisal}
The predictor is the V-JEPA~2 ViT-L model for 64-frame clips of $256 \times 256$ pixels~\cite{vjepa2}, run on a 4\,fps grid.
These frames are used only to score tokens.
The reader still receives the stream at 1\,fps, and no token from the additional frames enters the memory.
Each frame is resized to a short side of 292 pixels and center-cropped to $256 \times 256$, so the predictor sees the central part of a wide frame.
Each prediction window contains 56 context frames (14\,s) followed by 8 target frames (2\,s, four tubelets), aligned with the current block.
Each grid time takes the last native frame that has already arrived, so the context never contains a frame of the target block.
The channel scale $\sigma_{b,d}$ is computed from earlier blocks' channel-wise median absolute errors only.
It starts with a running mean and switches to an exponential average with update weight $1/30$ after 30 blocks with available prediction errors, giving a nominal time constant of 60\,s.
Scores are invalid during an 8\,s warm-up (4 blocks) after the first complete window, so the first scored block starts at 22\,s.
At the end of the warm-up, $\sigma_{\min}$ is fixed to the 10\% quantile of the channel scales.
Normalized channel errors are clipped at $\kappa=5$ before averaging.
Averaging patch scores over the four target tubelets yields the crop's $16 \times 16$ error map, which is pooled onto the reader's token grid by area overlap.
Tokens outside the crop receive no score.
They are kept by the coverage share, or by the even stride described below when a segment runs out of scored tokens.

\subsection{Online Budget Allocator}
\label{sec:supp_allocator}

\smallskip
\noindent\textbf{Segmentation.}
The surprisal-peak rule requires $\bar{u}_b > \bar{u}_{b-1}$ and $\bar{u}_b \ge \bar{u}_{b+1}$, so a peak is confirmed one block late; its robust $z$-score $\zeta_b$ uses the median and MAD of at least four valid blocks in the preceding 64\,s.
We use $\zeta_0 = 1.5$ and $U_{\max} = 4$ across benchmarks.
On the profiled OVO-Bench streams, segments last 7.8\,s on average.
The accumulation and duration rules close the open segment at the end of the latest block.
The minimum segment duration is 4\,s and the maximum is 16\,s; shorter segments are closed only by question arrival.
The boundary placed at the start of the anchor when a question arrives depends on when the question arrives, not on what it asks, so token-selection scores are still formed without the question.
A bounded buffer of open blocks lets the question-time boundary be placed without revising committed segments or knowing future query times.

\smallskip
\noindent\textbf{Retention schedule.}
At stream time $t$, a historical segment ending at $e_j$ has age $a_j=t-e_j$. Its retention ratio $r(a_j)$ decays online according to \cref{eq:retention}, with floor $\rho=0.3$ and timescale $T=20$\,s.
For $N_j$ original visual tokens, the resulting budget is $B_j(t)=\operatorname{round}(r(a_j)N_j)$. A question arriving at $t$ reads the allocation reached at that time. The anchor retains all its tokens.

\smallskip
\noindent\textbf{Allocation.}
We set the coverage share to $\alpha=0.5$, so a segment at retention ratio $r$ reserves $q_b=\operatorname{round}(rN_b/2)$ tokens of each block $b$ for coverage.
Coverage uses farthest-point sampling on the cosine similarity of the merged visual tokens that are stored.
With $\hat{x}_{b,i}=x_{b,i}/\lVert x_{b,i}\rVert$ and $\bar{x}_b$ the mean of the $\hat{x}_{b,i}$, the coverage order of block $b$ starts from
\begin{equation}
  \pi_1=\operatorname*{arg\,max}_{i}\ \hat{x}_{b,i}^{\top}\bar{x}_b
\end{equation}
and repeatedly adds the token whose highest similarity to the tokens already chosen is lowest,
\begin{equation}
  \pi_{m+1}=\operatorname*{arg\,min}_{i\notin\{\pi_1,\dots,\pi_m\}}\ \max_{l\le m}\ \hat{x}_{b,i}^{\top}\hat{x}_{b,\pi_l}.
\end{equation}
Ties go to the lowest index, so the order is deterministic.
The coverage set is the prefix $\mathcal{F}_b=\{\pi_1,\dots,\pi_{q_b}\}$.
Because $q_b$ never exceeds $\operatorname{round}(N_b/2)$, the order is computed once to that length, and any budget takes a prefix of it.
The surprisal share takes the remaining $B_j(t)-\sum_b q_b$ tokens of highest $u_{b,i}$ among the scored tokens of $S_j$ outside $\bigcup_b\mathcal{F}_b$, with no per-block quota.
If a block of the segment has no valid score, as during the warm-up, the whole surprisal share uses even-stride sampling instead, and the same sampling fills the rest when the scored tokens run out.
The even-stride budget is split across blocks by largest remainder in proportion to their unselected tokens.
A block that receives $c$ of its $n$ unselected tokens keeps those at positions $\lfloor (m+\tfrac12)\,n/c\rfloor$, $m=0,\dots,c-1$, in token order.

\smallskip
\noindent\textbf{Online pruning.}
Rounded coverage quotas and even-stride filling can produce non-nested selections, so a smaller budget may select a token that the current budget leaves out.
The online memory therefore retains a conservative superset of tokens that any later budget can select, and deletes tokens only when they leave this set.
Physical storage can temporarily exceed the current allocation $\widetilde{X}_j(t)$; once the minimum-retention quotas are reached, only that allocation remains.
Farthest-point orders and token surprisal rankings are reused throughout, and discarded visual content is never re-encoded.

\subsection{Evidence Gate}
\label{sec:supp_gate}
Segment keys use PE-Core-L14-336 frame embeddings sampled at 1\,fps.
Each segment has one key, the normalized mean of its frame embeddings.
For shortlisting, the text tower embeds the question and any supplied options in pieces of at most 30 tokens.
Relevance starts with the segment's highest cosine similarity to a piece of the question. For multiple-choice questions, we add 0.25 times its highest similarity to a piece of an option. We standardize relevance over eligible segments and set it to zero when fewer than three are eligible.
Shortlisting uses $k = 12$, $\beta = 1$, and $T_{\mathrm{r}}=60$\,s; admission uses a shared threshold $\tau = 2.2$ across readers and a cap of $M = 4$.
Candidates are scored in independent passes.
The anchor is processed once and shared as a prefix across them.
Specifically, the chat-template header and the anchor are prefilled once, and every pass continues from a copy of this cache, including the Gated DeltaNet states of Qwen3.8, so it processes only the candidate's tokens and the question.
A pass reads the Yes and No logits at the first answer position and generates no text, so its cost is that of a short prefill.
Because the passes are independent, our streaming implementation groups candidates of similar length into batches of up to four, bounded by 4,096 padded tokens, and restores their order afterwards.
Batching changes only the execution order, and each pass sees the same inputs as when run alone.
We did not tune this batch size; with more GPU memory, larger batches would score the shortlist in fewer sequential steps.

\subsection{Streaming Constraints and Reproducibility}
\label{sec:supp_protocol}

\smallskip
\noindent\textbf{Causality.}
The world model predicts each incoming block from preceding frames. When the block arrives, its observed features provide the prediction target, and the reader encodes the block once into native visual tokens.
The allocator reuses each block's coverage order and surprisal scores as the segment's retention ratio decays. Discarded visual tokens are never re-encoded.
Question arrival sets a boundary at the start of the anchor. This boundary depends on the arrival time, while segmentation scores and token selection are independent of the question text.
The resulting bank serves successive questions on the same stream, about 37 per hour of video on OVO-Bench.

\smallskip
\noindent\textbf{Reader compatibility.}
The same procedure runs on all four readers without training.
Surprisal is pooled onto each reader's own token grid. The Qwen readers merge two frames into one temporal patch, while LLaVA-OneVision encodes frames individually (\cref{sec:supp_visual}).
Because evidence scoring copies the cache of the anchor instead of storing caches for history, it also applies to hybrid models such as Qwen3.8, whose Gated DeltaNet layers keep fixed-size recurrent states rather than per-token keys and values.

\smallskip
\noindent\textbf{Determinism.}
\method answers by greedy decoding, and the evidence score is read from logits without sampling.
Farthest-point ties go to the lowest index, and batching changes only the execution order, so a run reproduces its answers up to floating-point differences across hardware.

\subsection{Prompts and Context Layout}
\label{sec:supp_prompts}
We use the reader's chat template. Each visual block carries its timestamp before its visual tokens. The timestamps in the examples below are illustrative.

\smallskip
\noindent\textbf{Evidence scoring.}
For each shortlisted segment, the reader sees the shared anchor first, followed by the candidate and the current question. Multiple-choice questions include the labeled options shown below. StreamBench uses the same evidence question with the options line omitted.
The gate reads $s_j=\ell_{\mathrm{Yes}}-\ell_{\mathrm{No}}$ at the first answer position without generating text.

\begin{promptbox}{Evidence scoring on a multiple-choice question}
The video right now:
<120.0 seconds><visual tokens> <122.0 seconds><visual tokens> ...
An earlier moment from the video's memory:
<64.0 seconds><visual tokens> <66.0 seconds><visual tokens> ...
A question is being asked about a video that is still playing. "The video right now" shows the frames of the present moment. "An earlier moment from the video's memory" shows frames that were recorded earlier in the same video.

Question: {question}
Options: A. {option}; B. {option}; C. {option}; ...

Does the earlier moment provide additional visual evidence that is needed to answer this question, beyond what the video right now already shows? Evidence for only a part of the question also counts. Answer Yes or No.
\end{promptbox}

\smallskip
\noindent\textbf{Answering and scoring multiple-choice questions.}
For OVO-Bench, StreamingBench, Video-MME, MLVU, and LongVideoBench, we present the benchmark question and its original options. The scorer extracts the predicted choice from the response and compares it with the annotated choice. \Cref{sec:supp_benchmarks} gives the aggregation for each benchmark.
When history is admitted, its segments appear in chronological order before the anchor. The answer prompt follows the visual input. On StreamingBench, we use the SimpleStream answer prompt~\cite{simplestreaming}, with the benchmark's supplied dialogue context for sequential questions.

\begin{promptbox}{StreamingBench answer with admitted history}
Earlier moments from the video's memory:
<36.0 seconds><visual tokens> ... <42.0 seconds><visual tokens>
<64.0 seconds><visual tokens> ... <70.0 seconds><visual tokens>
The video right now:
<120.0 seconds><visual tokens> ... <126.0 seconds><visual tokens>
{multiple-choice task instructions}
Question: {question}
Options:
A. {option A}
B. {option B}
C. {option C}
D. {option D}
Only give the best option's letter (A, B, C, or D) directly.
\end{promptbox}

\smallskip
\noindent\textbf{Answering and scoring StreamBench questions.}
On StreamBench~\cite{streamchat}, the reader receives the current open-ended question with any earlier questions and its own generated answers from the same dialogue. It answers in free text.
Either answer format can use admitted history or the anchor alone. When the anchor is the only visual input, both visual labels are omitted, as in the example below.

\begin{promptbox}{StreamBench answer from the anchor alone}
<120.0 seconds><visual tokens> ... <126.0 seconds><visual tokens>
Answer the question about the video concisely.
Previous conversation (earlier turns in this video):
User: {earlier question}
Assistant: {reader's earlier answer}
Current question: {question}
\end{promptbox}

\noindent Following the released StreamBench evaluation, a Llama~3 8B Instruct judge receives the question, the generated answer, and the reference answer. It judges whether the two answers agree in meaning, allowing synonyms and paraphrases. Reported accuracy is the percentage of Yes judgments. For both answer formats, \method decodes greedily for up to 128 tokens.

\section{Limitations and Broader Impact}
\label{sec:supp_impact}

\subsection{Limitations}
Deploying \method requires GPU memory for two auxiliary models alongside the frozen reader. V-JEPA~2 and the PE-Core retrieval encoder together add about 3.7\,GiB of fixed weights. With Qwen3.8-27B, the weights of all three models occupy 54.64\,GiB across two GPUs before the stream begins (\cref{sec:supp_efficiency}). The visual memory bank resides in host memory. Across the four readers we evaluated, \method stores and reuses native visual tokens and reads the first-step Yes and No logits to score evidence (\cref{sec:supp_gate}). Integrating a reader through a hosted API that returns only generated text would require support for these operations.

\subsection{Broader Impact}
\method builds a query-agnostic visual memory by selecting visual tokens before it knows what a viewer will ask. The reader encodes each incoming block once and, under the default settings, inspects at most twelve historical segments per question. With host storage for the growing memory bank, this design can in principle process a live stream without a predetermined end time (\cref{sec:supp_cost}). For example, a viewer joining a live workshop hours after it began could ask which tool the instructor used in an earlier step. The same memory could also let someone ask later about a bystander who briefly appeared on camera. In shared spaces, a deployment should make recording visible to people in view, restrict who can query the history, and give users control over when it is deleted.

\end{document}